\pdfoutput=1

\documentclass{article}

\usepackage{microtype}
\usepackage{graphicx}
\usepackage{subfigure}
\usepackage{booktabs} 

\usepackage{hyperref}

\usepackage[accepted]{icml2024}

\usepackage{amsmath}
\usepackage{amssymb}
\usepackage{mathtools}
\usepackage{amsthm}
\usepackage{makecell}

\usepackage[capitalize,noabbrev]{cleveref}

\theoremstyle{plain}

\theoremstyle{definition}

\theoremstyle{remark}

\usepackage[textsize=tiny]{todonotes}

\usepackage[utf8]{inputenc}
\usepackage[small]{caption}
\usepackage{booktabs}
\usepackage[switch]{lineno}
\usepackage{xcolor}
\usepackage[linesnumbered,ruled,vlined]{algorithm2e}
\usepackage{float}

\usepackage{wrapfig}
\usepackage{pifont}

\usepackage{amssymb}
\usepackage{amsmath,amsfonts}
\usepackage{amsopn}
\usepackage{bm} 
\usepackage{multirow}
\usepackage{tabularx}
\newcommand{\Ours}{\method{FixBN}\xspace}
\newcommand{\BN}{\method{BN}\xspace}
\newcommand{\GN}{\method{GN}\xspace}
\newcommand{\WN}{\method{WN}\xspace}
\newcommand{\LN}{\method{LN}\xspace}
\newcommand{\IN}{\method{IN}\xspace}

\newcommand{\Fixup}{\method{Fixup}\xspace}

\newcommand{\FedAvg}{\method{FedAvg}\xspace}

\newcommand{\FedTAN}{\method{FedTAN}\xspace}

\newcommand{\FedDNA}{\method{FedDNA}\xspace}
\newcommand{\FedBN}{\method{FedBN}\xspace}

\newcommand{\HeteroFL}{\method{HeteroFL}\xspace}

\newcommand{\vct}[1]{\boldsymbol{#1}} 

\newcommand{\ProbOpr}[1]{\mathbb{#1}}

\newcommand{\expect}[2]{%
\ifthenelse{\equal{#2}{}}{\ProbOpr{E}_{#1}}
{\ifthenelse{\equal{#1}{}}{\ProbOpr{E}\left[#2\right]}{\ProbOpr{E}_{#1}\left[#2\right]}}} 

\newcommand{\vgamma}{\vct{\gamma}}
\newcommand{\vbeta}{\vct{\beta}}
\newcommand{\vtheta}{\vct{\theta}}

\newcommand{\vmu}{\vct{\mu}}
\newcommand{\vsigma}{\vct{\sigma}}

\newcommand{\vS}{\vct{S}}
\newcommand{\vh}{\vct{h}}

\newcommand{\vx}{{\vct{x}}}

\newcommand{\sB}{\mathcal{B}}

\newcommand{\sD}{\mathcal{D}}

\newcommand{\sL}{\mathcal{L}}

\newcommand{\eat}[1]{}
\newcommand{\method}[1]{\textsc{#1}}

\SetKwInput{KwInput}{Input}                
\SetKwInput{KwOutput}{Output}              
\SetKwInput{KwInit}{Initialize}              

\newcommand{\cmark}{\color{green}{\ding{51}}}%
\newcommand{\xmark}{\color{red}{\ding{55}}}%

\usepackage{colortbl}
\definecolor{ForestGreen}{HTML}{009B55}
\definecolor{RoyalPurple}{HTML}{613F99}

\definecolor{Gray}{gray}{0.9}
\definecolor{LightCyan}{rgb}{0.88,1,1}
\renewcommand{\paragraph}[1]{\vspace{-0.5ex}\textbf{#1}}
\newcommand{\ie}{i.e.\xspace}
\newcommand{\eg}{e.g.\xspace}

\usepackage{enumitem}

\icmltitlerunning{Making Batch Normalization Great in Federated Deep Learning}

\begin{document}

\twocolumn[
\icmltitle{Making Batch Normalization Great in Federated Deep Learning}

\icmlsetsymbol{equal}{*}

\begin{icmlauthorlist}
\icmlauthor{Jike Zhong}{equal,yyy}
\icmlauthor{Hong-You Chen}{equal,yyy}
\icmlauthor{Wei-Lun Chao}{yyy}

\end{icmlauthorlist}

\icmlaffiliation{yyy}{Department of Computer Science and Engineering,  The Ohio State University, Columbus, USA}

\icmlcorrespondingauthor{Wei-Lun Chao}{chao.209@osu.edu}

\icmlkeywords{Machine Learning, ICML}

\vskip 0.3in
]

\printAffiliationsAndNotice{\icmlEqualContribution} 

\begin{abstract}

Batch Normalization (\BN) is widely used in {centralized} deep learning to improve convergence and generalization.
However, in {federated} learning (FL) with decentralized data, 
prior work has observed that training with \BN could hinder performance and suggested replacing it with Group Normalization (\GN).
In this paper, we revisit this substitution by expanding the empirical study conducted in prior work. Surprisingly, we find that \BN outperforms \GN in many FL settings. The exceptions are high-frequency communication and extreme non-IID regimes. We reinvestigate
factors that are believed to cause this problem, including the mismatch of \BN statistics across clients and the deviation of gradients during local training. We empirically identify a simple practice that could reduce the impacts of these factors while maintaining the strength of \BN. Our approach, which we named \Ours, is fairly easy to implement, without any additional training or communication costs, and performs favorably across a wide range of FL settings. We hope that our study could serve as a valuable reference for future practical usage and theoretical analysis in FL.

\end{abstract}

\section{Introduction}
\label{s-intro}

Federated learning (FL) is a decentralized optimization framework in which several clients collaborate, usually through a server, to achieve a common learning goal without exchanging data \cite{kairouz2019advances,li2020federated-survey}. FL has attracted a lot of attention lately due to the increasing concern about data privacy, protection, and ownership. 
The core challenge is how to obtain a machine learning model whose performance is as good as if it were trained in a conventional centralized setting, especially under the practical condition where the data across clients are non-IID.

For models normally trained via stochastic gradient descent (SGD), such as deep neural networks (DNNs), \FedAvg \cite{mcmahan2017communication} is the most widely used FL training algorithm.
\FedAvg iterates between two steps: parallel local SGD at the clients, and global model aggregation at the server. 
In the extreme case where global aggregation takes place after every local SGD step (\ie, high communication frequency), \FedAvg is very much equivalent to centralized SGD for training simple DNN models like multi-layer perception \cite{zhou2017convergence,stich2019local,haddadpour2019convergence,li2020convergence,zhao2018federated}.

\begin{figure}
    \centering
    \includegraphics[width=0.75\linewidth]{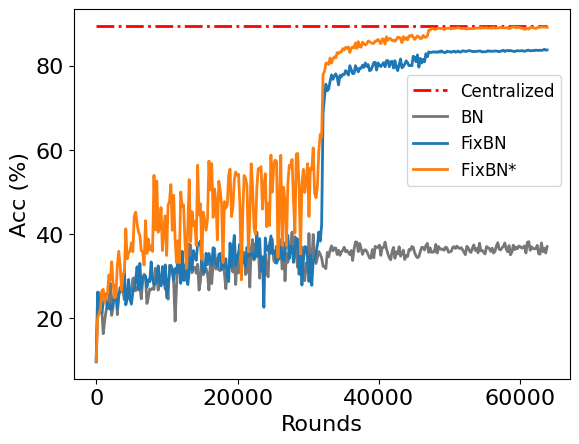}
    \vskip -7pt
    \caption{\small \textbf{Our approach, \Ours, notably bridges the gap of using \BN in FL and centralized learning.} X-axis: communication rounds in FL, after every local SGD step; y-axis: test accuracy on CIFAR-10~\cite{krizhevsky2009learning}; $\star$: further with our SGD momentum in FL. Please see~\autoref{sec:analysis} and~\autoref{s_mom} for details.}
    \label{fig:summary} 
    \vskip -12pt
\end{figure}

In this paper, we focus on DNN models that contain Batch Normalization (\BN) layers~\cite{ioffe2015batch}. In centralized learning, especially for deep feed-forward models like ResNet~\cite{he2016deep}, \BN has been widely used to improve the stability of training and speed up convergence. However, in the literature on FL, many of the previous experiments have focused on shallow ConvNets (CNN) without \BN. Only a few papers have particularly studied the use of \BN in FL~\cite{hsieh2020non,du2022rethinking,wang2023batch}. 
In the very first study, \citet{hsieh2020non} pointed out 
the mismatch between the feature statistics (\ie, means and variances in \BN) estimated on non-IID local mini-batches (in training) and global data (in testing), and argued that it would degrade \FedAvg's performance. 
\citet{hsieh2020non} thus proposed to replace \BN with Group Normalization (\GN)~\cite{wu2018group}, which does not rely on mini-batch statistics for normalization, and showed its superior performance in some FL settings. Such a solution has since been followed by a long non-exhaustive list of later work \cite{jin2022accelerated, charles2021large,lin2020ensemble,yuan2021we,reddi2020adaptive,hyeon2021fedpara,yu2021fed2,hosseini2021federated}.

That said, replacing \BN with \GN in FL seems more like an ad hoc solution rather than a cure-all. First, in centralized training, \BN typically outperforms \GN empirically. 
Second, several recent papers \cite{mohamad2022fedos,tenison2022gradient,yang2022towards,chen2021fedbe} have reported that \BN is still better than \GN in their specific FL applications. Third, changing the normalization layer may create a barrier between the communities of centralized learning and FL. To illustrate, in centralized learning, many publicly available pre-trained checkpoints~\cite{pytorch_hub,onnx_model} are based on popular CNN architectures with \BN; most understanding~\cite{bjorck2018understanding,santurkar2018does,luo2018towards}, empirical studies~\cite{garbin2020dropout}, and theoretical analysis~\cite{yang2019mean} about normalization in DNNs are built upon \BN. These prior results may become hard to be referred to in the FL community.

Motivated by these aspects, we revisit the problem of \BN in FL and the replacement of \BN with \GN.

To begin with, we conduct an extensive empirical study comparing \BN and \GN in FL on commonly used image classification datasets including CIFAR-10 \cite{krizhevsky2009learning} and Tiny-ImageNet \cite{le2015tiny}.
Specifically, we extend the study in \cite{hsieh2020non} by considering a wide range of non-IID degrees and communication frequencies. 
We find that \GN does not outperform \BN in many of the settings, especially when the communication frequency is low or the non-IID degree is not severe. This suggests that one should consider the FL setting when selecting the normalization method. 

Building upon this observation, we take a deeper look at the scenarios where \BN performs particularly poorly: high communication frequencies and severe non-IID degrees. We revisit the factors identified by \citet{hsieh2020non,du2022rethinking,wang2023batch} that are believed to impact \BN's effectiveness in FL. Specifically, \citet{wang2023batch} theoretically showed that the mismatched mini-batch statistics across non-IID clients lead to deviated gradients on local copies of the DNN models. What is worse, such an impact cannot be canceled even if clients communicate right after \emph{every} local SGD step, as also evidenced in \cite{wang2021quantized,zheng2020design,chai2021fedat}. 
We note that this does not happen to FL with \GN, as \GN does not use mini-batch statistics to normalize intermediate features.

This theoretical result seems to imply a dilemma of the use of \BN in FL. 
On the one hand, \BN relies on stochastic mini-batch statistics to normalize intermediate features to claim superior convergence and generalization~\cite{luo2018towards,santurkar2018does}. On the other hand, the \emph{dependency of the gradients on the non-IID local mini-batch} prevents \FedAvg from recovering the gradients computed in centralized learning, even under high-frequency communication.

{Can we resolve, or at least, alleviate, this dilemma?} We investigate the \emph{training dynamics} of FL with \BN. 
We find that even with the deviated gradients, each local copy of the DNN model converges regarding its local training loss. 
Moreover, the variation of mini-batch statistics \emph{within} each client reduces along with the communication rounds. 
These observations imply that the positive impact of \BN diminishes in the later rounds, opening up the option of ceasing the use of mini-batch statistics for normalization to reduce the negative impact on FL due to the mismatch across clients.
 
Taking this insight into account, we propose a simple yet effective practice named \Ours, which requires no architecture change, no additional training, and no extra communication. \Ours starts with the standard practice of \BN in \FedAvg: in local training, using mini-batch statistics for normalization while accumulating local \BN statistics; in global aggregation,  
aggregating the accumulated local \BN statistics into global ones.
Then after a decent amount of communication rounds, \Ours freezes the \BN layer while keeping other DNN parameters learnable. Namely, in this phase, local training does not rely on the mini-batch statistics but the frozen accumulated global statistics to normalize features, which \emph{(a)} allows \FedAvg to recover the centralized gradient under high communication frequency settings and \emph{(b)} removes the mismatch of normalization statistics in training and testing. As shown in~\autoref{fig:summary}, \Ours improves \BN by a large margin in the high-frequency regime. We further demonstrate \Ours's effectiveness on extensive FL settings, outperforming or on a par with \GN and \BN.

Together with our study and the proposed approach, we also reveal several interesting observations of \BN, suggesting the need for a deeper investigation and theoretical analysis of \BN in FL. We hope that our study could serve as a valuable reference for future research directions and practice in FL.

\section{Related Work}
\label{sec:related}
\paragraph{Federated learning.} Many methods have been proposed to improve \FedAvg~\cite{mcmahan2017communication} from different perspectives like server aggregation~\cite{wang2020federated,yurochkin2019bayesian,lin2020ensemble,he2020group,zhou2020distilled,chen2021fedbe}, server momentum~\cite{hsu2019measuring,reddi2021adaptive}, local training \cite{yuan2020federated,liang2019variance,li2019feddane,feddyn,li2020federated,karimireddy2020scaffold}, pre-training~\cite{chen2022pre,nguyen2023begin}, etc. Many papers in FL have also gone beyond training a single global model. Personalized FL~\cite{kairouz2019advances} aims to create models that are tailored to individual clients, \ie, to perform well on their data distributions. It is achieved through techniques such as federated multi-task learning~\cite{li2020federated-FMTL,smith2017federated}, model interpolation~\cite{mansour2020three}, and fine-tuning~\cite{yu2020salvaging,chen2021bridging}. In this paper, \emph{we investigate the impact of the \BN layers in DNNs in standard (\textbf{not} personalized) FL.}

\paragraph{Normalization layers in centralized training.}
Since \BN was introduced~\cite{ioffe2015batch}, normalization layers become the cornerstone of DNN architectures. The benefits of \BN have been extensively studied in centralized training such as less internal covariate shift~\cite{ioffe2015batch}, smoother optimization landscape~\cite{santurkar2018does}, robustness to hyperparameters~\cite{bjorck2018understanding} and initialization~\cite{yang2019mean}, accelerating convergence~\cite{ioffe2015batch}, etc. The noise of the estimated statistics of BN in mini-batch training is considered a regularizer~\cite{luo2018towards} that improves generalization~\cite{ioffe2015batch}. A recent study~\cite{lubana2021beyond} showed that \BN is still the irreplaceable normalizer vs. a wide range of choices in general settings. Unlike in FL, \emph{\BN often outperforms \GN in standard centralized training.}

\paragraph{Existing use of normalizers in FL.}
In the context of FL, less attention has been paid to normalization layers. \citet{hsieh2020non} were the first to suggest replacing \BN with \GN for non-IID decentralized learning. Several work~\cite{du2022rethinking,zhang2023normalization} reported that \LN is competitive to \GN in FL. \citet{hong2021federated} enhanced adversarial robustness by using statistics from reliable clients but not for improving performance. \citet{diao2020heterofl} proposed to normalize batch activations instead of tracking running statistics for the scenario that the clients have heterogeneous model architectures. 
These papers propose to \emph{replace} \BN while we aim to \emph{reclaim \BN's superiority} in FL. 

Several work~\cite{duan2021feddna,idrissi2021fedbs} proposed dedicated server aggregation methods for \BN statistics 
(separated from other model parameters) for specific tasks. 
For multi-modal learning, \citet{bernecker2022fednorm} proposed maintaining a different \BN layer for each modality instead of sharing a single one. In personalized FL, \citet{li2021fedbn,andreux2020siloed,jiang2021tsmobn} proposed to maintain each client's independent \BN layer, inspired by the practice of domain adaptation in centralized training~\cite{li2016revisiting}. \citet{lu2022personalized} leveraged \BN statistics to guide aggregation for personalization. We note that the goals of these papers are orthogonal to ours. 

The most related to ours is \cite{wang2023batch}, which provided a theoretical analysis of \BN in FL and proposed a modified use of \BN, \FedTAN, to reduce divergent gradients in local training. However, \FedTAN requires additional communication rounds linear to the numbers of \BN layers, which is much more expensive than
our \Ours.
\section{Background}
\label{sec:analysis}

\paragraph{Batch Normalization (\BN).} 
The \BN layer is widely used as a building block in feed-forward DNNs. Given an input feature vector $\vh$, it normalizes the feature via the mean $\vmu_{\sB}$ and variance $\vsigma^2_{\sB}$ computed on a batch of features $\sB$, followed by a learnable affine transformation (via $\vgamma, \vbeta$): 
\begin{align}
\hat\vh = f_{\BN}(\vh; (\vgamma, \vbeta), (\vmu_{\sB}, \vsigma^2_{\sB}))=\vgamma\frac{\vh-\vmu_{\sB}}{\sqrt{\vsigma^2_{\sB} + \epsilon}} + \vbeta; \label{eq:bn}\\
\epsilon\ \:\text{is a small constant}.
\nonumber
\end{align}
In training, the statistics $\vmu_{\sB}$ and $\vsigma^2_{\sB}$ are computed on each training mini-batch during the forward passes. Meanwhile, these mini-batch statistics are accumulated by the following exponential moving average (controlled by $\alpha$):
\begin{align}
&\vmu:=\alpha\vmu+(1-\alpha)\vmu_{\sB}, \quad\vsigma^2:=\alpha\vsigma^2+(1-\alpha)\vsigma^2_{\sB}.
\label{eq:bn_stat}
\end{align}
During testing, $\vmu$ and $\vsigma^2$ are seen as the estimated statistics of the whole training set and used for normalization, \ie, to replace $\vmu_{\sB}$ and $\vsigma^2_{\sB}$
in \autoref{eq:bn}.

\paragraph{Federated Learning (FL).}
In a federated setting, the goal is to learn a model on the training data distributed among $M$ clients. Each has a training set $\sD_m = \{(\vx_i, y_i)\}_{i=1}^{|\sD_m|}, \forall m\in[M]$, where $\vx$ is the input (\eg, images) and $y$ is the true label. Let $\sD = \cup_m \sD_m$ be the aggregated training set from all clients; $\ell$ is the loss function on a data sample. FL aims to minimize the empirical risk over all the clients: 
\begin{align}
& \min_{\vtheta}~\sL(\vtheta) = \sum_{m=1}^M \frac{|\sD_m|}{|\sD|} \sL_m(\vtheta), \hspace{10pt}\\
& \text{where} \hspace{10pt} \sL_m(\vtheta) = \frac{1}{|\sD_m|} \sum_{i=1}^{|\sD_m|} \ell(\vx_i, y_i; \vtheta),
\label{eq:obj}
\end{align}
where $\vtheta$ is the model parameters. For DNNs with \BN layers, $\vtheta$ includes learnable weights of the perceptron layers like fully-connected and convolutional layers, in addition to the statistics $\{(\vgamma, \vbeta), \vS\}$ of all \BN layers, where $\vS=(\vmu, {\vsigma}^2)$ are the accumulate \BN means and variances. 

\begin{figure*}[t]
    \centering
    \minipage{1\linewidth}
    \minipage{0.5\linewidth}
    \centering
    \vskip-5pt
    \includegraphics[width=1\linewidth]{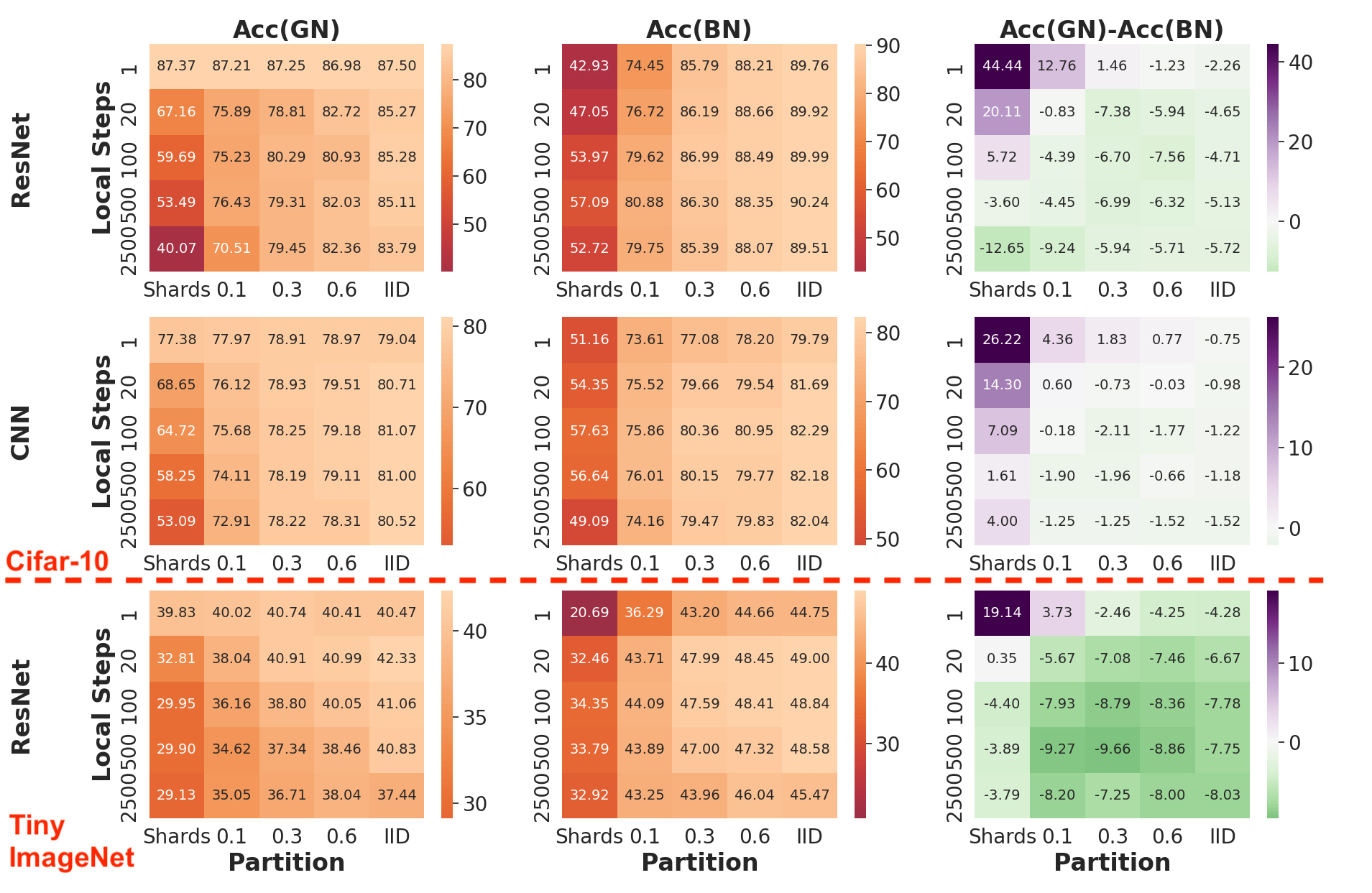} 
    \mbox{\small (a) Fixed 128 \textbf{epochs}}
    \endminipage
    \hfill
    \minipage{0.5\linewidth}
    \centering
    \vskip-5pt
    \includegraphics[width=1\linewidth]{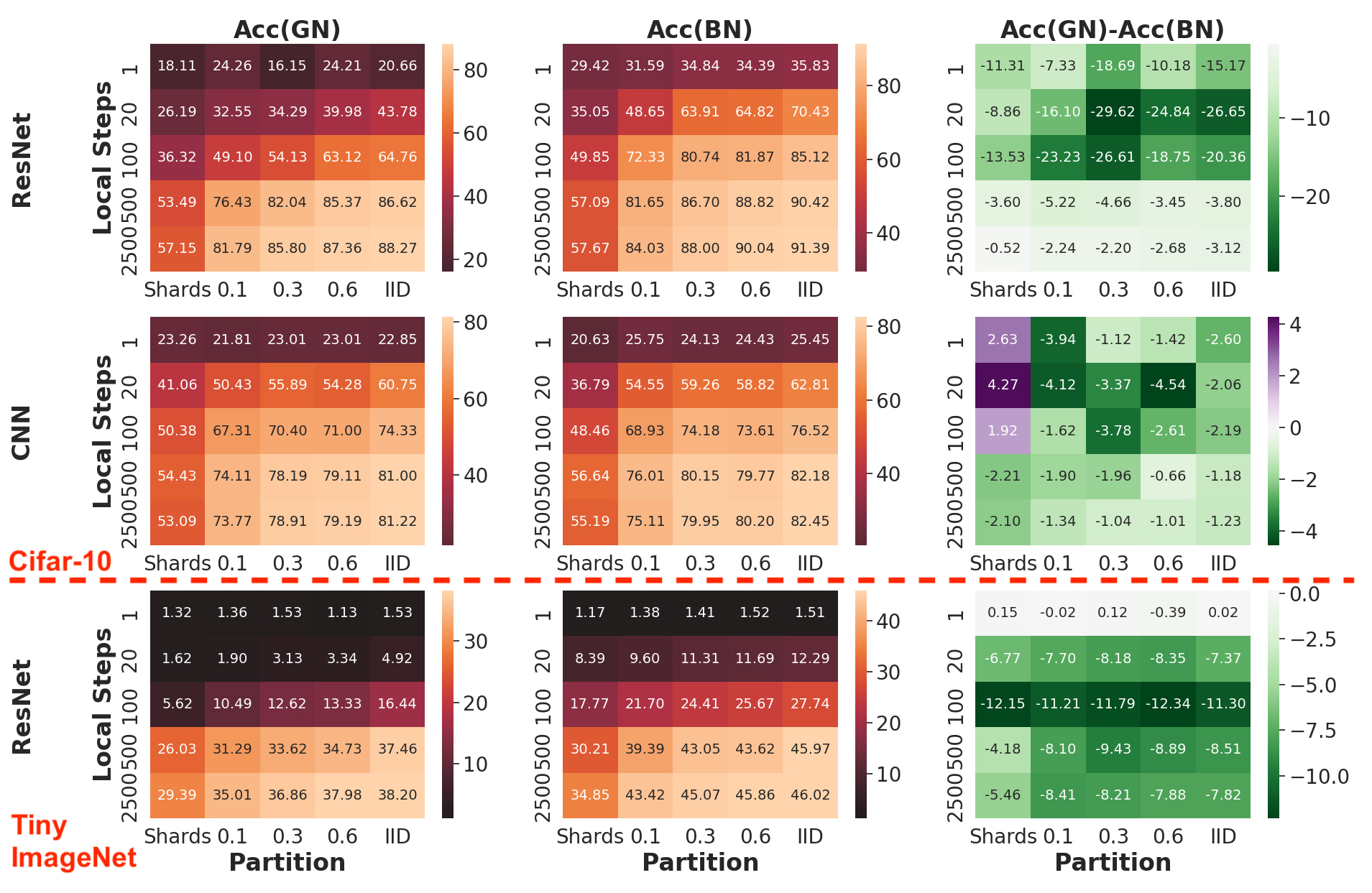}
    \mbox{\small (b) Fixed 128 \textbf{rounds}}
    \endminipage
    \endminipage
    \vskip -5pt
    \caption{\small \textbf{Is \GN consistently better than \BN in FL? No.} We compare their test accuracy in various FL settings on CIAFR-10 and Tiny-ImageNet, including different non-IID partitions and numbers of local steps $E$. We consider \textbf{(a)} a fixed budget of the total number of SGD steps (\eg, for CIFAR-10,  $20\text{ local steps } \times 20\text{ batch size }\times 5\text{ clients }\times 3200\text{ rounds} = 128\text{ epochs of CIFAR-10 training data}$) or \textbf{(b)} a fixed number of total communication rounds ($128$ rounds). {\color{ForestGreen}Green} cells: \BN outperforms \GN. {\color{RoyalPurple}Purple} cells: \GN outperforms \BN.} 
    \label{fig:GNBN}
    \vskip -10pt
\end{figure*}

\paragraph{Federated averaging (\FedAvg).} \autoref{eq:obj} can not be solved directly in an FL setting due to the decentralized data. The fundamental FL algorithm \FedAvg~\cite{mcmahan2017communication} solves \autoref{eq:obj} by multiple rounds of parallel local updates at the clients and global model aggregation at the server. Given an initial model $\bar{\vtheta}^{(0)}$, for round $t=1, ..., T$, \FedAvg performs:
\begin{align}
& \textbf{Local:  } \vtheta_m^{(t)} = \texttt{ClientUpdate}(\sL_m, \bar{\vtheta}^{(t-1)}); \\
& \textbf{Global:  } \bar{\vtheta}^{(t)} \leftarrow \sum_{m=1}^M \frac{|\sD_m|}{|\sD|}{\vtheta_m^{(t)}}.
\label{eq_global}
\end{align}
During local training, the clients update the model parameters received from the server, typically by minimizing each client's empirical risk $\sL_m$ with several steps (denoted as $E$) of mini-batch SGD. For the locally accumulated means and variances in \BN, they are updated by~\autoref{eq:bn_stat}. During global aggregation, all the parameters in the locally updated models $\{\vtheta_m^{(t)}\}$, including \BN statistics, are averaged element-wise over clients. Typically, $E\gg 1$ due to communication constraints.

\paragraph{Group Normalization (\GN) as a replacement in FL.} 
The main challenge in \FedAvg is that the distributions of local data $\{\sD_m\}$ can be drastically different across clients. \citet{hsieh2020non,du2022rethinking} argued that this non-IID issue is particularly problematic for DNNs with \BN layers since they depend on the activation means and variances estimated on non-IID mini-batches. 
A popular alternative suggested by~\cite{hsieh2020non} is \GN \cite{wu2018group}. 
GN divides the adjacent channels into groups and normalizes within each group. It operates separately on each data instance, thus removing
the issue in BN.
This solution is followed by many later papers as mentioned in~\autoref{s-intro} and~\autoref{sec:related}.

\section{Expanded Empirical Study on \BN vs.~\GN}
\label{sec:study_new}

Despite the widespread replacement of \BN with \GN in FL, several papers have reported that \BN performs better than \GN in their applications~\cite{mohamad2022fedos,tenison2022gradient,yang2022towards,chen2021fedbe}. \emph{After checking~\cite{hsieh2020non} in detail, we find that their empirical study only considers extreme non-IID settings (\ie, each client's data comes from a subset of the label space, a.k.a.~the Shards setting) with a specific number of local steps $E$ (\eg, $E=20$ for CIFAR-10).} This motivates us to expand their study with a wider range of FL settings.

\paragraph{Experiment setup.}
We use the standard FL benchmark CIFAR-10 \cite{krizhevsky2009learning} and Tiny-ImageNet \cite{le2015tiny}. 
We consider different FL setting including \textbf{(1) non-IID degrees}, ordered in increasing skewness: {IID, Dirichlet({0.1, 0.3, 0.6}), and Shards}. We follow \cite{hsu2019measuring} for Dirichlet sampling. For Shards, each client only has data from $20\%$ of the classes. We consider 5/10 clients for CIFAR-10/Tiny-ImageNet, respectively, similarly to~\cite{hsieh2020non}. As practical FL is constrained on computation and communication, we consider two \textbf{(2) budget criteria}: {fixed 128 epochs} of total local SGD updates over all the clients and communication rounds, and {fixed 128 rounds} of total communication. Within every round, we consider different \textbf{(3) local steps}: $E\in\{1, 20, 100, 500, 2500\}$.

We use ResNet20~\cite{he2016deep} and LeNet-like CNN~\cite{lecun1998gradient} for CIFAR-10 and ResNet18 for Tiny-ImageNet. For training hyperparameters, we generally follow~\cite{hsieh2020non} to use the SGD optimizer with $0.9$ momentum, learning rate $0.02$ (decay by $0.1$ at $50\%$ and $75\%$ of the total rounds, respectively), batch size $20$, and full participation for default unless specified. The group size of \GN is tuned and set at $2$. \emph{The average of $3$ runs of \FedAvg is reported.} 
{Please see the appendix for more details, results, discussions, and full reproducibility.}

\paragraph{Results.} 
We summarize the test accuracy with \BN and \GN on CIFAR-10 and Tiny-ImageNet in~\autoref{fig:GNBN}, across different FL settings.
We highlight the following observations: 
\begin{itemize}[nosep,topsep=0pt,parsep=0pt,partopsep=0pt, leftmargin=*]
\item \textbf{No definite winner.} \GN is often considered the default replacement for \BN in previous FL work. 
However, according to \autoref{fig:GNBN}, \GN is not always better than \BN. 
\item \textbf{\BN outperforms \GN in many settings.} 
This can be seen from the {\color{ForestGreen} green} cells in ``Acc(GN)-Acc(BN)'' heatmaps. Specifically, when clients cannot communicate frequently (\ie, many local steps like $E=100\sim2500$), \BN seems to be the better choice for normalization. 
\item \textbf{\GN outperforms \BN in extreme cases.} 
We find that \GN outperforms \BN 
(the {\color{RoyalPurple}purple} cells in ``Acc(GN)-Acc(BN)'' heatmaps) only in extreme non-IID (\eg, Shards) and high frequency (\eg, $E=1~\sim 20$) settings. This aligns with the reported settings in~\cite{hsieh2020non}.
\item \textbf{Consistency over more settings.} We verify in the appendix that factors like participation rates and the number of total clients do not change the above observations.
\item \textbf{\BN cannot recover centralized performance.} We dig deeper into the extremely high frequency setting at $E=1$; \ie, clients communicate after every local SGD step.
At first glance, this should recover mini-batch SGD in centralized learning (\eg, training on multi-GPUs with local shuffling). 
However, as shown in~\autoref{tbl:centralized} and~\autoref{fig:summary}, there is a huge accuracy gap (about $45\%$) between centralized and federated learning for DNNs with \BN. As a reference, \emph{such a gap very much disappears for DNNs with \GN}. We investigate this finding further in~\autoref{ss_problem}.
\end{itemize}

\begin{table}
\tabcolsep 2pt
\caption{\small \textbf{BN cannot recover centralized performance with communication after \emph{every step} ($E=1$).} We train a ResNet20 with either \BN or \GN on non-IID CIFAR-10 (5 clients, Shards). Both the FL and centralized training use SGD without momentum.}
\label{tab:small}
\vskip-8pt
\centering
\footnotesize
\renewcommand{\arraystretch}{0.75}
\begin{tabular}{c|cc}
\toprule
Norm & Centralized Acc. & FL Acc.\\
\midrule
\GN & 87.46$\pm$0.57& 87.37$\pm$1.16\\
\BN & 89.30$\pm$0.89& {\color{red}42.93}$\pm$2.75 \\
\bottomrule
\end{tabular}
\label{tbl:centralized}
\vskip-10pt
\end{table}

\paragraph{Remark.} The goal of this empirical study is to provide a more complete picture of whether one should replace \BN with \GN, not to answer all the questions regarding their performance. 
Even in centralized learning, a full understanding of the comparison between \BN and \GN is still lacking, not to mention the more complicated federated learning. 
We also want to emphasize that our empirical result is not against existing work. In contrast, it unifies both ends of the existing findings: \citet{hsieh2020non} showed that \GN outperforms \BN in extreme non-IID and high-frequency settings while \citet{mohamad2022fedos,tenison2022gradient,yang2022towards,chen2021fedbe} showed that \BN is still preferred in their specific (low frequency) settings.

Looking at each normalized method separately, we note that the accuracy change of \GN follows the common understanding of \FedAvg: the accuracy degrades with severe non-IID settings and many local steps (under the fixed epochs setting). For \BN, its accuracy also degrades with severe non-IID settings, which aligns with the arguments in \cite{hsieh2020non,du2022rethinking,wang2023batch}. Interestingly, beyond our expectations, \BN performs fairly robustly along with the increase of local steps and even benefits from it. This can be seen in \autoref{fig:GNBN} (a) ACC (BN) column-wise. We provide further analysis in~\autoref{sec:study}. 
\section{Rethinking Batch Normalization in FL} 
\label{ss_problem}

Seeing that \BN outperforms \GN in many FL settings, we dig deeper into its poorly performing regimes: high communication frequencies and severe non-IID degrees.

\subsection{\BN makes the gradients biased in local training} 
\label{ss:bn_gradient}
As pointed out by~\citet{wang2023batch} in their theoretical analysis, the mismatch of mini-batch statistics across non-IID clients leads to deviated local gradients, which cannot be canceled out even under high-frequency settings. 
As a simple illustration, we derive the \textbf{F}orward-\textbf{B}ackward pass of the plain \BN layer $f_{\BN}$ (see~\autoref{eq:bn}) for one input example $\vx_i$ in a mini-batch $\sB$.
\begin{align}
&\textbf{F:  }\ell(\hat \vx_i) = \ell(f_{\BN}(\vx_i; (\vgamma, \vbeta), {(\vmu_{\sB}, \vsigma^2_{\sB})})) \label{eq_bn_forward}\\ & \hspace{33.5pt}=\ell(\vgamma\frac{\vx_i-\vmu_{\sB}}{\sqrt{\vsigma^2_{\sB} + \epsilon}} + \vbeta)=\ell(\vgamma\tilde{\vx_i} + \vbeta);\nonumber\\ 
&\textbf{B:  }    \frac{\partial{\ell}}{\partial{\vx_i}} \hspace{3.5pt}= \label{eq_bn_backward} 
\frac{|\sB|\frac{\partial{\ell}}{\partial{{\color{red}\tilde{\vx_i}}}} -\sum_{j=1}^{|\sB|}{\frac{\partial{\ell}}{\partial{{\color{red}\tilde{\vx}_{j}}}}} -
{\color{red}\tilde{\vx_i}}\hspace{2pt}{\color{black}\cdot}\hspace{2pt}
\sum_{j=1}^{|\sB|}{\frac{\partial{\ell}}{\partial{{\color{red}\tilde{\vx}_{j}}}}\hspace{2pt}{\color{black}\cdot}\hspace{2pt}{{\color{red}\tilde{\vx}_{j}}}}}{|\sB|\sqrt{{\color{red}\vsigma_{\sB}^{2}}+\epsilon}},
\end{align}
where $\ell$ is an arbitrary loss function on the \BN layer's output $\hat \vx_i$, ``$\cdot$'' is element-wise multiplication, and $\frac{\partial{\ell}}{\partial{{\color{red}\tilde{\vx}}}}=\vgamma\frac{\partial{\ell}}{\partial{\hat \vx}}$.

We can see that many terms in~\autoref{eq_bn_backward} (colored in {\color{red}red})  depend on the mini-batch features $\{\vx_j\}_{j=1}^{|\sB|}$ or mini-batch statistics ${(\vmu_{\sB}, \vsigma^2_{\sB})}$. The background gradient $\frac{\partial{\ell}}{\partial{\vx_i}}$ w.r.t.~the input vector $\vx_i$ is thus sensitive to what other examples in the mini-batch are. 
Suppose $\vx_i$ belongs to client $m$ in a non-IID FL setting, the gradient $\frac{\partial{\ell}}{\partial{\vx_i}}$ is doomed to be different when (a) it is calculated locally with other data sampled from $\sD_m$ and when (b) it is calculated globally (in centralized learning) with other data sampled from $\sD = \cup_m \sD_m$. Namely, even if clients communicate after every
mini-match SGD step, \emph{how a particular data example influences the DNN parameters is already
different between FL and centralized training.}
This explains why in \autoref{tbl:centralized}, \BN cannot recover the centralized learning
performance.

\begin{figure*}[t]
    \centering
    \minipage{1\linewidth}
    \minipage{0.25\linewidth}
    \centering
    \vskip-5pt
    \includegraphics[width=1\linewidth]{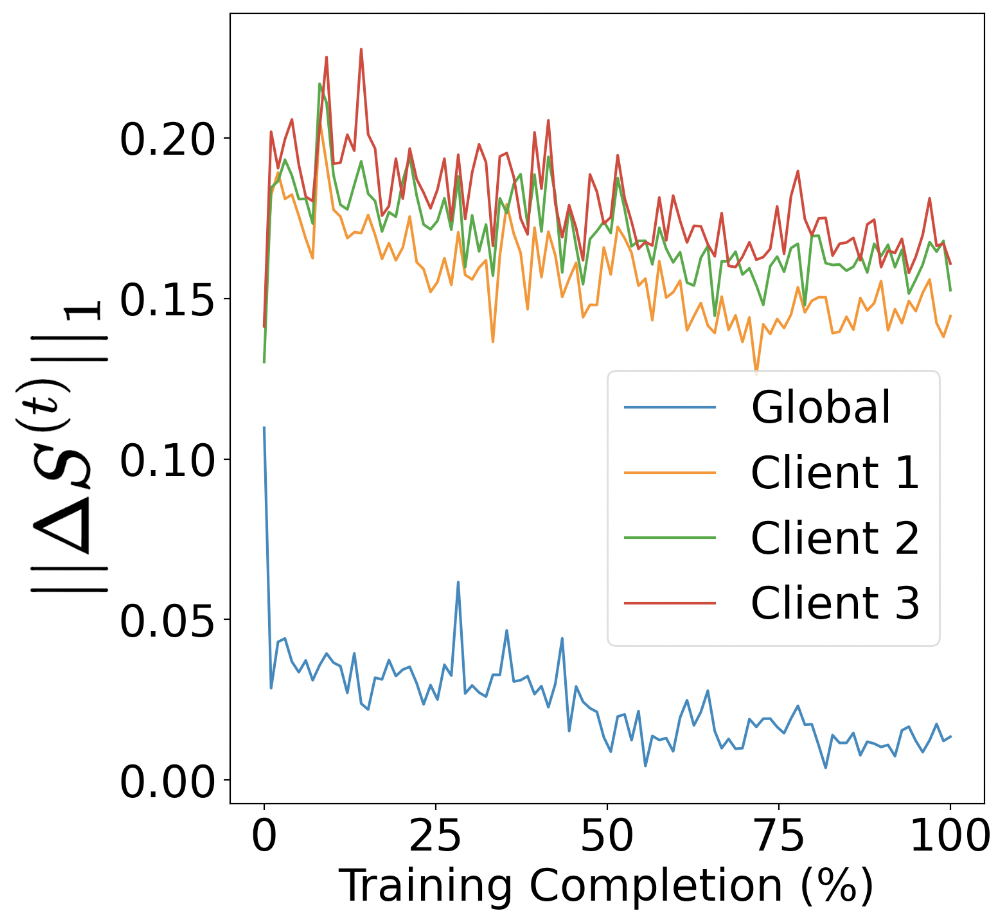} 
    \mbox{\small (a)}
    \endminipage
    \hfill
    \minipage{0.25\linewidth}
    \centering
    \vskip-5pt
    \includegraphics[width=1\linewidth]{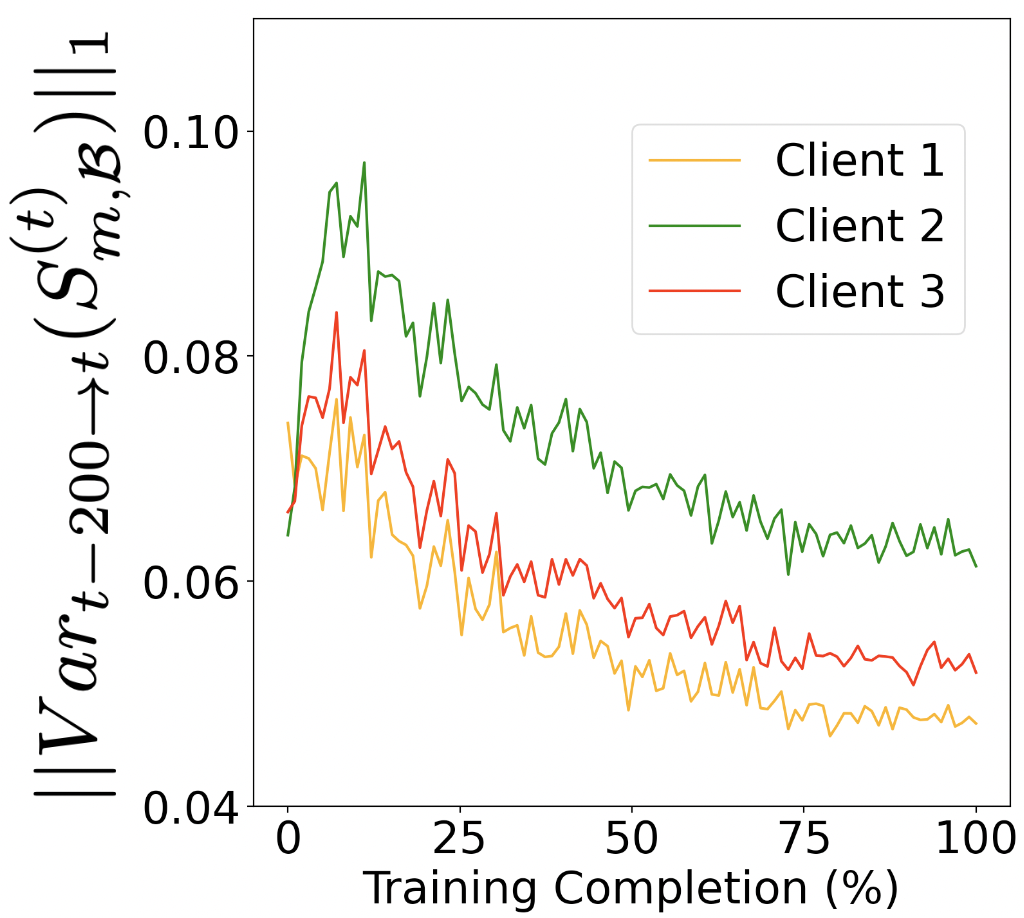}
    \mbox{\small (b)}
    \endminipage
    \hfill
    \minipage{0.25\linewidth}
    \centering
    \vskip-5pt
    \includegraphics[width=1\linewidth]{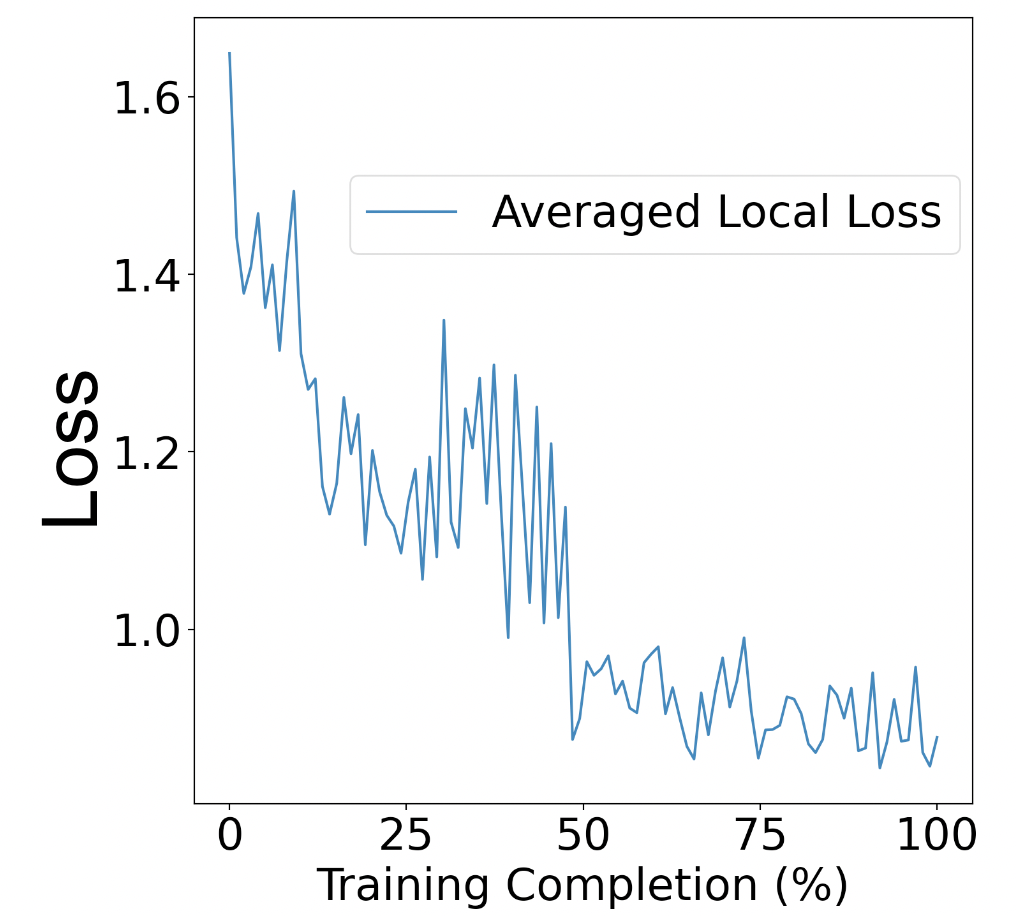}
    \mbox{\small (c)}
    \endminipage
    \hfill
    \minipage{0.25\linewidth}
    \centering
    \vskip-5pt
    \includegraphics[width=1\linewidth]{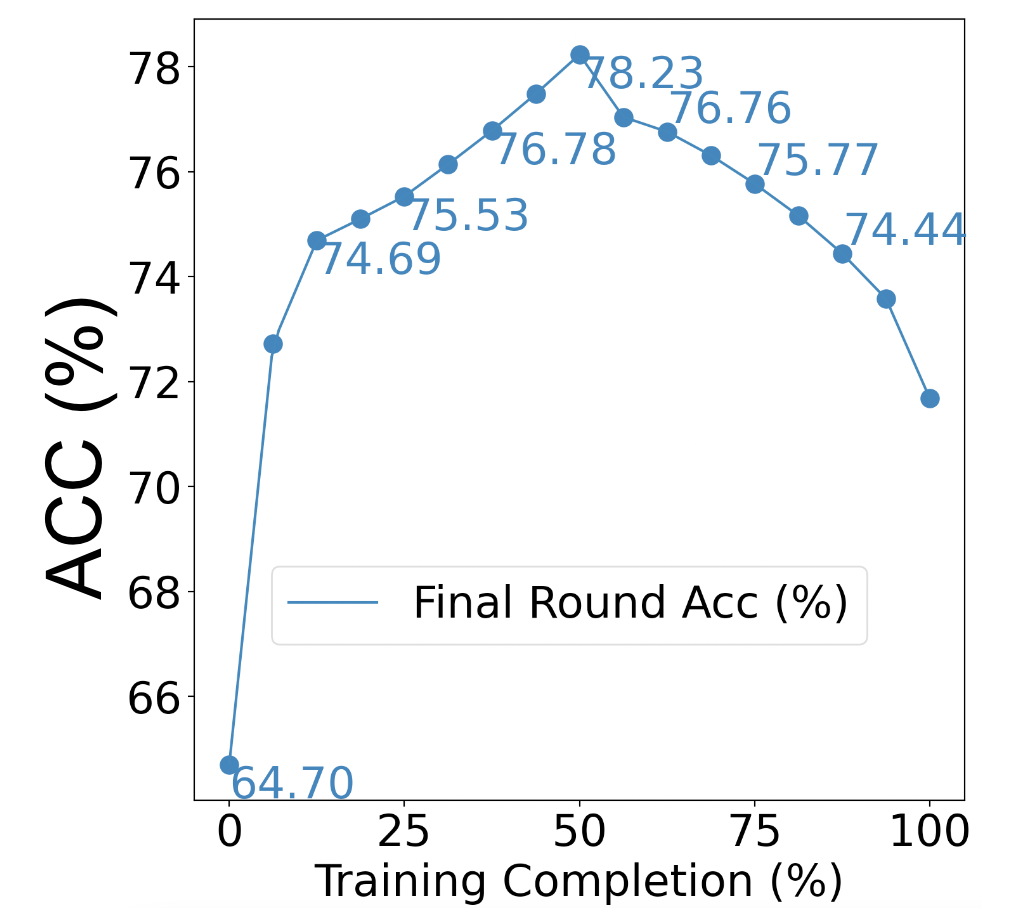}
    \mbox{\small (d)}
    \endminipage
    \vskip -10pt
    \endminipage
    \vskip -10pt
    \caption{\small \textbf{Training dynamics of \FedAvg with \BN.} (a) Changes of global accumulated statistics ($\|\bar{\vS}^{(t+1)}-\bar\vS^{(t)}\|_1$) and deviations of local mini-batch statistics from global accumulated statistics ($\|{\vS_{m, \sB}}^{(t+1)}-\bar\vS^{(t)}\|_1$). (b) Variances (running over $t-200$ to $t$) of local mini-batch statistics $\vS_{m, \sB}^{(t)}$. (c) Averaged local training losses over clients. (d) Final-round accuracy when freezing \BN statistics at different intermediate rounds in a non-IID CIFAR-10 setting
    (shards, $E$ = 100) using ResNet20. See the main text for details.} 
    \label{fig:fixbn_motiv}
    \vskip -10pt
\end{figure*}

\subsection{Mismatched normalization in training and testing}
\label{ss:mismatch}
Besides the deviated gradients, another factor impeding \BN's performance in a non-IID FL setting is the mismatched normalization statistics between training and testing \cite{hsieh2020non}. As reviewed in~\autoref{sec:analysis}, \BN uses mini-batch statistics for normalization in training and accumulated statistics for normalization in testing. In \FedAvg, the former is calculated from local data in local training; the latter is aggregated from locally accumulated statistics and applied to the global model in testing. In a non-IID setting, the former is doomed to deviate from the latter, creating a non-trivial discrepancy between training and testing even in the later rounds of \FedAvg.

\subsection{The dilemma of applying \BN in FL}
\label{ss:dilemma}
Despite the negative impacts on \FedAvg, mini-batch statistics are the key to \BN's positive impacts on DNN training (see~\autoref{sec:related}). Specifically, normalization with mini-batch statistics makes DNN training robust to internal covariate shifts and gradient explosion~\cite{lubana2021beyond,yang2019mean,ioffe2015batch}, especially in the early training stages when the model weights and intermediate activations are changing rapidly. The stochastic nature of mini-batch statistics also facilitates the search for a flatter loss landscape~\cite{luo2018towards,santurkar2018does}.

\section{\Ours: Towards Breaking the Dilemma of \BN in Federated Learning}
\label{sec:method}

\subsection{Insights from the training dynamics}
\label{ss_insight}
\emph{Is there an opportunity to alleviate \BN's negative impacts on FL while maintaining its positive impacts on DNN training?} 
We start our exploration by taking a deeper look at the training dynamics of DNN with \BN in standard \FedAvg. Under the same $E=1$ setting as in~\autoref{tbl:centralized} unless stated otherwise, we highlight four critical observations.

\begin{enumerate}[nosep,topsep=0pt,parsep=0pt,partopsep=0pt, leftmargin=*]
\item As shown in \autoref{fig:fixbn_motiv} (a), the local mini-batch statistics ${\vS_{m, \sB}}^{(t)}$ estimated from client $m$ during round $t$\footnote{As we set $E=1$, within each communication round, we only perform one mini-batch SGD.} remain largely different from the global 
accumulated statistics $\bar\vS^{(t)}$, even at later \FedAvg rounds. This discrepancy results from the non-IID local data.
While not surprisingly, the result re-emphasizes the potentially huge negative impacts of the factors in~\autoref{ss:bn_gradient} and \autoref{ss:mismatch}.

\item Still in \autoref{fig:fixbn_motiv} (a), we find that the global accumulated statistics (in {\color{blue}blue}) gradually converge. 

\item As shown in \autoref{fig:fixbn_motiv} (b), the variation of the local mini-batch statistics ${\vS_{m, \sB}}^{(t)}$ within each client reduces along with the \FedAvg training rounds. 

\item As shown in \autoref{fig:fixbn_motiv} (c), even with deviated local gradients, the averaged training loss $\sum_{m=1}^M \frac{|\sD_m|}{|\sD|} \sL_m(\vtheta_m^{(t)})$ (in {\color{blue}blue}) of local models $\{\vtheta_m^{(t)}\}_{m=1}^M$ converges.
\end{enumerate}
Putting these observations together, we find that the positive impacts of \BN diminish along with \FedAvg training, as evidenced by the reduced variation of mini-batch statistics (point 3) and the convergence of the local models (point 4).
This opens up the opportunity to cease the use of local mini-batch statistics for normalization in later rounds, to resolve the deviated gradient issue~\autoref{ss:bn_gradient}. 
Specifically, as the global accumulated statistics converge, it makes sense to freeze and use them for normalization, which would resolve the mismatched statistics issue in \autoref{ss:mismatch}.

We examine this idea by replacing local mini-batch statistics with fixed global statistics starting at different rounds. As shown in~\autoref{fig:fixbn_motiv} (d), if the round is chosen properly (x-axis), the corresponding \emph{final round test accuracy} 
largely improves (y-axis). Based on this insight, we propose a simple practice to address the issues of BN mentioned  in~\autoref{ss_problem}.

\subsection{\Ours by two-stage training}
\label{ss_two_stage}
The insights in~\autoref{ss_insight} motivate us to divide \FedAvg training with \BN into two stages, separated at round $T^\star$. 

\paragraph{Stage I: Exploration (before round $T^\star$).} In this stage, we follow the standard practice of \FedAvg with \BN. 
In each SGD step during local training, we calculate the mini-batch statistics in the forward pass to normalize the intermediate features (\autoref{eq_bn_forward}) and calculate the gradients for model updates in the backward pass (\autoref{eq_bn_backward}). Meanwhile, we accumulate the local mini-batch statistics using \autoref{eq:bn_stat}. During global aggregation, we perform element-wise averages for all the DNN parameters across clients, including \BN's accumulated statistics. At the next round of local training, the local accumulated statistics are initialized with the globally aggregated accumulated statistics. 

This stage enjoys the positive impacts of \BN on local training with SGD, \eg, robust to internal covariate
shifts when the learning rate is large and model weights and intermediate activations are fast changing. 

At the end, we save the aggregated accumulated statistics $\bar\vS^{(T^\star)}$ of each \BN layer from the global model $\bar{\vtheta}^{(T^\star)}$.

\begin{figure}[t]
    \centering
    \minipage{0.47\linewidth}
    \centering
    \includegraphics[width=1\linewidth]{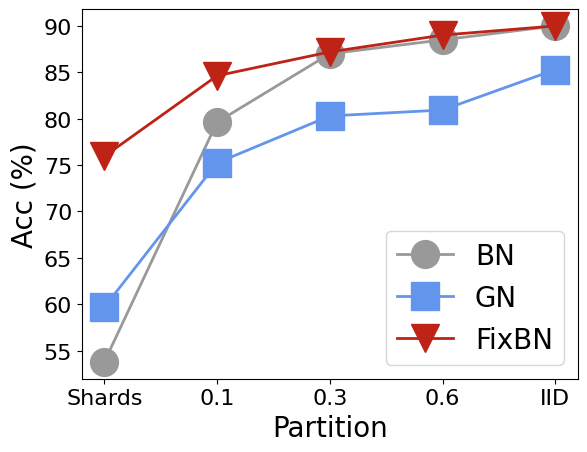}
    \vskip -5pt
    \mbox{\small (a) CIFAR-10}
    \endminipage
    \hfill
    \minipage{0.47\linewidth}
    \centering
    \includegraphics[width=1\linewidth]{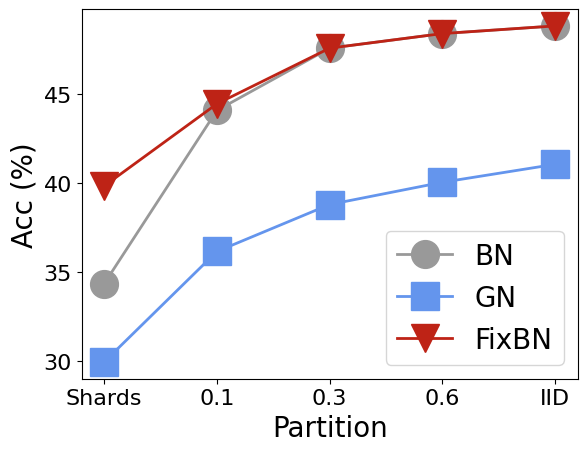} 
    \vskip -5pt
    \mbox{\small (b)  Tiny-ImageNet}    
    \endminipage
    \vspace{-5pt}
    \caption{\small Non-IID partitions with $E=100$ steps.}
    \vspace{-15pt}
    \label{fig:fix_fix}
\end{figure}

\paragraph{Stage II: Calibration (after round $T^\star$).} 
After the DNN parameters have benefited from \BN in the early stage of training, we seek to address the negative impacts of \BN on \FedAvg. As mentioned in \autoref{ss:bn_gradient} and \autoref{ss:mismatch}, these are caused by normalization with mismatched mini-batch statistics across non-IID clients. In the second stage, we thus replace mini-batch statistics with the \emph{frozen} global accumulated statistics $\bar\vS^{(T^\star)}$ to normalize activations in local training. We implement this idea by switching \BN from the training mode to the evaluation mode during local training. We note that with this implementation, the global statistics $\bar\vS^{(T^\star)}$ at round $T^\star$ will also be used during testing (see~\autoref{sec:analysis}), essentially removing the mismatched statistics in training and testing (\autoref{ss:mismatch}). Methods for selecting $T^\star$ is discussed in \autoref{suppl-sec: indicator}.

\paragraph{Remark.} We use $\bar\vS^{(T^\star)}$ to approximate the global statistics because $\bar\vS^{(t)}$ converges in the later rounds of \FedAvg (see \autoref{fig:fixbn_motiv} (a)). 
In terms of $T^\star$, we find it quite stable to be set at the middle round and fix it at $50\%$ of the total \FedAvg rounds throughout our experiments, unless stated otherwise.
\Ours is easy to apply, \emph{with no architecture and objective change and no extra training and communication costs}. Please see \autoref{suppl-sec:algo} for the pseudo-code of \Ours. 

\section{Maintaining Local SGD Momentum}
\label{s_mom}
Besides \BN, we identify another gap between \FedAvg and centralized training. While it is common to use SGD momentum during local training of \FedAvg, the momentum is usually discarded at the end of each round and re-initialized (along with any optimizer states) at the beginning of the next round of local training. That is, the first several SGD steps in a round cannot benefit from it. 

We present a simple treatment: keeping the \textbf{local momentum} without re-initialization at the end of each local training round. This makes \FedAvg a stateful method. 
A stateless choice is to maintain the \textbf{global momentum}~\cite{karimireddy2020mime} by uploading the local momentum to the server after each round and aggregating it by~\autoref{eq_global}. This global momentum is then sent back to clients to initialize the momentum of the next local training round, with the cost of double message size. Empirically, we find the two methods yield similar gains (\autoref{fig:fix_mom}) and help recover centralized accuracy if communicating every step (\autoref{fig:summary}).

\section{Experiments (more in the \autoref{suppl-sec:exp_r})}
\label{ss_fixbn_exp}

\begin{table}[t]
\newcommand{\specialcell}[2][c]{%
  \begin{tabular}[#1]{@{}c@{}}#2\end{tabular}}
\caption{\small Dirichlet (0.1) non-IID results on ImageNet-1K.}
\vskip -10pt
\label{tab:small}
\centering
\footnotesize
\setlength{\tabcolsep}{1.25pt}
\renewcommand{\arraystretch}{0.75}
\begin{tabular}{l|c|l}
\toprule
Method & Network & Acc. $\color{magenta}\Delta_{\text{-BN}}$\\
\midrule
\GN & \multirow{3}{*}{\specialcell{ResNet18\\\cite{he2016deep}}} & 33.33\scriptsize$\pm$ 0.57\\
\BN & & 48.30\scriptsize$\pm$ 1.21 \\
\textbf{\Ours} & & \textbf{52.43}\scriptsize$\pm$ 0.68 \textcolor{magenta}{(+4.1)}\\
\bottomrule
\end{tabular}
\vskip -10pt
\label{tbl:imgnet}
\end{table}

\begin{table}
\newcommand{\specialcell}[2][c]{%
  \begin{tabular}[#1]{@{}c@{}}#2\end{tabular}}
\caption{\small Mean IoU ($\%$) of image segmentation on Cityscapes.}
\vskip -10pt
\label{tab:small}
\centering
\footnotesize
\setlength{\tabcolsep}{2.25pt}
\renewcommand{\arraystretch}{0.75}
\begin{tabular}{l|c|l}
\toprule
Method & Backbone & Mean IoU  $\color{magenta}\Delta_{\text{-BN}}$\\
\midrule
\GN & \multirow{3}{*}{\specialcell{MobileNet-v2\\\cite{sandler2018mobilenetv2}}} & 43.2{\scriptsize$\pm$0.33} \\
\BN & & 48.9{\scriptsize$\pm$0.36}\\
\textbf{\Ours} & & \textbf{54.0}{\scriptsize$\pm$0.29} \textcolor{magenta}{(+5.1)}\\
\midrule
\GN & \multirow{3}{*}{\specialcell{ResNet50\\\cite{he2016deep}}} & 47.8{\scriptsize$\pm$0.30} \\
\BN & & 52.6{\scriptsize$\pm$0.38}\\
\textbf{\Ours} & & \textbf{57.2}{\scriptsize$\pm$0.32} \textcolor{magenta}{(+4.6)}\\
\bottomrule
\end{tabular}
\label{tbl:cityscapes}
\vskip-10pt
\end{table}

\subsection{Main results}
We first compare our \Ours (w/o momentum) to \GN and \BN.
The group size of \GN is tuned and set to $2$~\cite{hsieh2020non}. The average of $3$ runs of \FedAvg is reported.

\paragraph{Results on CIFAR-10 and Tiny-ImageNet.} We follow the setups in~\autoref{sec:study_new} unless stated otherwise. For CIFAR-10, we use ResNet20~\cite{he2016deep}; for Tiny-ImageNet, we use ResNet18. We consider $E=100$ and run \FedAvg for a \textbf{fixed 128 epochs} of total local SGD updates over all the clients and communication rounds. 
\autoref{fig:fix_fix} shows the results. \Ours consistently outperforms \BN and \GN, especially in severe non-IID cases.

\paragraph{Results on ImageNet.} We extend \Ours to the ImageNet-1K~\cite{liimagenet} dataset, which is split into 100 non-IID clients with Dirichlet (0.1) over classes. We learn a ResNet18 with 10\% randomly sampled clients per round, 20 batch size, 0.1 learning rate (decay by 0.1 every 30\% of the total rounds), $E=2$ local epochs, and 64 epochs in total. \autoref{tbl:imgnet} shows that \Ours also perform the best.  

\paragraph{Results on realistic non-IID Cityscape.}
We further consider a natural non-IID setting on the image segmentation dataset Cityscape~\cite{Cordts2015Cvprw}. We treat each ``city'' as a client and train $100$ \FedAvg rounds using DeepLab-v3+~\cite{chen2018encoder}. More details are in the appendix. \autoref{tbl:cityscapes} shows that \Ours's effectiveness is generalized to different architectures and vision tasks.

\subsection{More analysis}

\label{ss:ind}
\begin{figure}
    \vskip 0pt
    \hfill
    \minipage{1\linewidth}
·   \minipage{1\linewidth}
    \centering
    \includegraphics[width=0.75\linewidth]{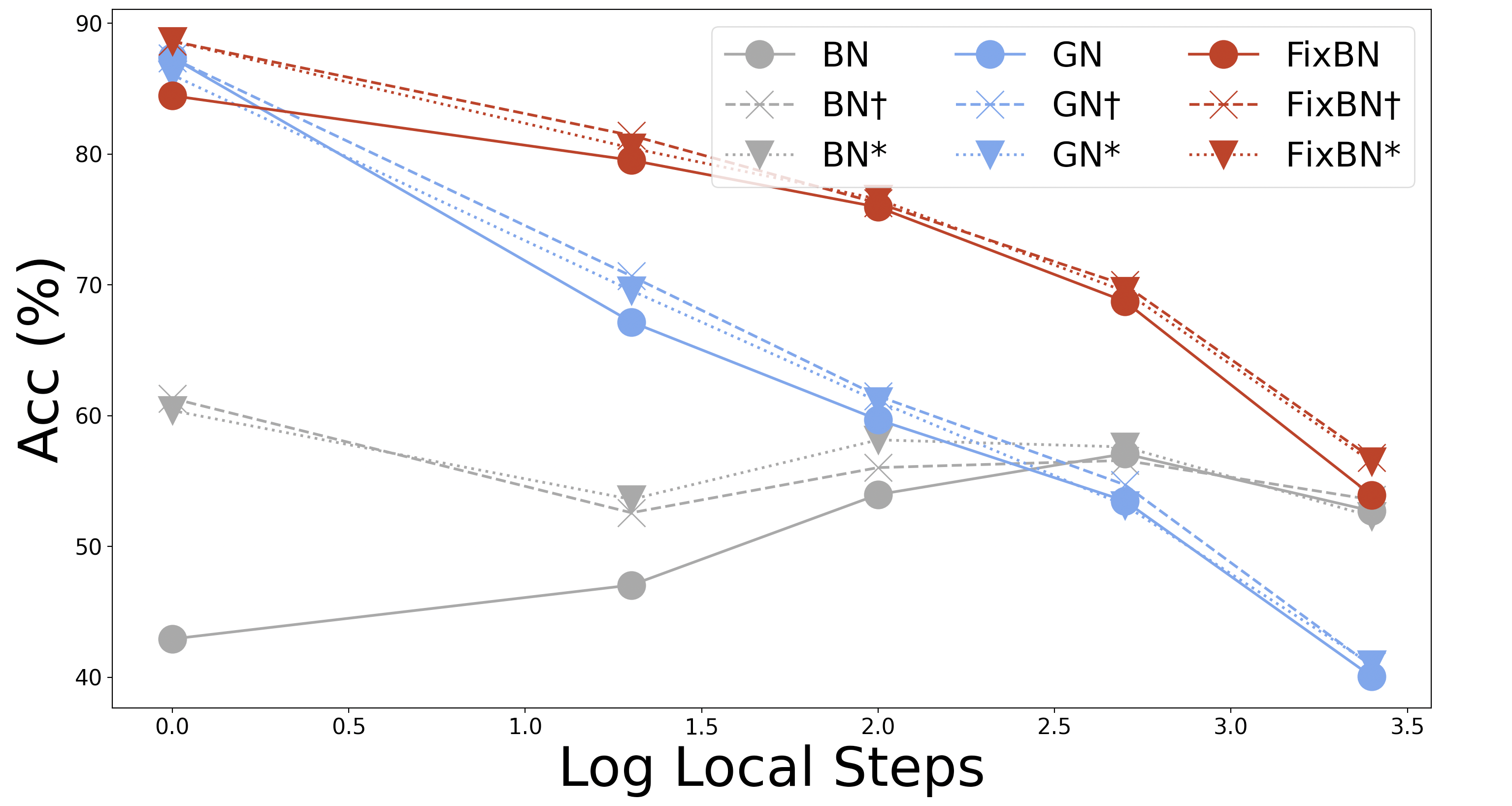}
    \endminipage
    \endminipage
    \vspace{-8pt}
    \caption{\small \textbf{Maintained momentum results.} We apply maintained \textbf{global ($^\dagger$)} and \textbf{local momentum ($^\star$)} to \FedAvg with different normalizers. The setting is (Shards, fixed 128 epochs) with different local steps $E$. We show results on CIFAR-10. Results on  Tiny-ImageNet follow similar trends (see \autoref{fig:fix_mom_tiny} in appendix).}
    \vspace{-20pt}
    \label{fig:fix_mom}
\end{figure}

\paragraph{Maintained SGD momentum.}
We apply the maintained local momentum and global momentum proposed in~\autoref{s_mom} to \FedAvg with different normalizers. We take the Shards setting with fixed 128 epochs and consider different numbers of local steps per round: $E\in\{1, 20, 100, 500, 2500\}$. As shown in~\autoref{fig:fix_mom}, the maintained SGD momentum consistently improves\FedAvg; the gains by global and local momentum are similar, providing flexibility for stateless and stateful use cases. More gains are at smaller $E$, supporting our motivation to fix the zero initialization issue. 

\paragraph{Performance at high-frequency regimes.} Still in~\autoref{fig:fix_mom} with a (Shards, fixed epoch) setting, we compare \Ours to \BN and \GN. \Ours performs consistently better. Importantly, \Ours remains highly accurate in high-frequency communication (\ie, small numbers of local steps $E$), unlike \BN, suggesting its effectiveness in mitigating the deviation and mismatched issues in~\autoref{ss_problem}.
Indeed, in~\autoref{fig:summary} with $E = 1$, \Ours largely recovers
centralized performance.

\paragraph{Other normalization methods.}
We consider other normalization layers in~\autoref{tbl:fixup}. \Ours performs favorably. Interestingly, the normalization-free \Fixup\cite{zhang2019fixup} initialization for residual networks\footnote{\citet{zhuang2023normalization} also reported improvement by scaled weight normalization, similar to~\cite{qiao2019micro} in~\autoref{tbl:fixup}.} notably outperforms \GN, suggesting a new alternative in FL besides \Ours.

In~\autoref{tbl:fedtan}, we further apply \Ours to the experimental setting in Table 1 of \FedTAN~\cite{wang2023batch} to compare to other \BN variants for FL. We note that \FedTAN needs $\Theta(3L+1)$ communication rounds linear to the numbers of \BN layers $L$, which is much more expensive than \Ours. \HeteroFL~\cite{diao2020heterofl} directly normalizes the activations, which cannot resolve the non-IID issue.   
We note that \FedBN~\cite{li2021fedbn} was proposed for personalized FL, beyond the scope of this paper.

\section{Further Analysis following~\autoref{sec:study_new}}
\label{sec:study}

We extend our discussion about~\autoref{fig:GNBN}. It is well-known that increasing the number of local steps per round leads to greater drifts of the model from centralized learning~\cite{karimireddy2020scaffold}. 
However, using more local steps also updates the local models more, potentially leading to an improved average model. To balance these competing considerations, we discuss two criteria. For \textbf{(a) fixed 128 epochs}, more local steps mean fewer communication rounds, in which \GN degrades monotonically ``as expected''. \emph{Interestingly, \BN has an opposite trend.} \BN actually improves and outperforms \GN with larger $E$s. For \textbf{(b) fixed 128 rounds}, understandably, using more local steps improves both \GN and \BN, since more local SGD updates are made in total. 

\begin{table}[t]
\caption{\small Normalization layers on CIFAR-10 (Shards, $E=100$).}
\label{tab:small}
\centering
\footnotesize
\vskip -7pt
\setlength{\tabcolsep}{1pt}
\renewcommand{\arraystretch}{0.75}
\begin{tabular}{l|c}
\toprule
Normalization Layer & Acc (\%)\\
\midrule
\BN\cite{ioffe2015batch} & 53.97 \scriptsize$\pm$ 4.18\\
\GN\cite{wu2018group} & 59.69 \scriptsize$\pm$ 0.76\\
\GN+\WN~\cite{qiao2019micro} & 66.90 \scriptsize$\pm$ 0.81\\ 
\LN\cite{ba2016layer} & 54.54 \scriptsize$\pm$ 1.21\\
\IN\cite{ulyanov2016instance} & 59.76 \scriptsize$\pm$ 0.43\\
\Fixup\cite{zhang2019fixup} & 70.66 \scriptsize$\pm$ 0.24\\
\midrule
\textbf{\Ours (Ours)}& \textbf{76.56} \scriptsize$\pm$ 0.66\\
\bottomrule
\end{tabular}
\label{tbl:fixup}
\vskip -5pt
\end{table}

\begin{table}[t]
\caption{\small \textbf{Other FL normalization methods.} The setting follows \cite{wang2023batch}: ResNet20 on CIFAR-10 given \# of rounds.}
\vskip-7pt
\label{tab:small}
\centering
\footnotesize
\setlength{\tabcolsep}{0.75pt}
\renewcommand{\arraystretch}{0.75}
\begin{tabular}{l|c|cc}
\toprule
FL Scheme & \#R & IID & Non-IID \\
\midrule
Centralized+\BN & - & \multicolumn{2}{c}{91.53}\\
Centralized+\Ours & - & \multicolumn{2}{c}{91.62}\\
\midrule
\FedTAN~\cite{wang2023batch} & 580K & 91.26 & 87.66\\
\FedAvg +\BN & 10K & 91.35&45.96\\
\FedAvg +\GN & 10K & 91.26&82.66\\
\HeteroFL~\cite{diao2020heterofl} & 10K &91.21 & 30.62\\
\FedDNA~\cite{duan2021feddna} & 10K &91.42 & 76.01\\
\midrule
\FedAvg+\Ours (Ours)& 10K & 91.35 &87.71\\
\bottomrule
\end{tabular}
\label{tbl:fedtan}
\vskip -10pt
\end{table}

\begin{figure}[h]
    \centering
    \vskip-5pt
    \minipage{0.5\linewidth}
    \centering
    \includegraphics[width=1\linewidth]{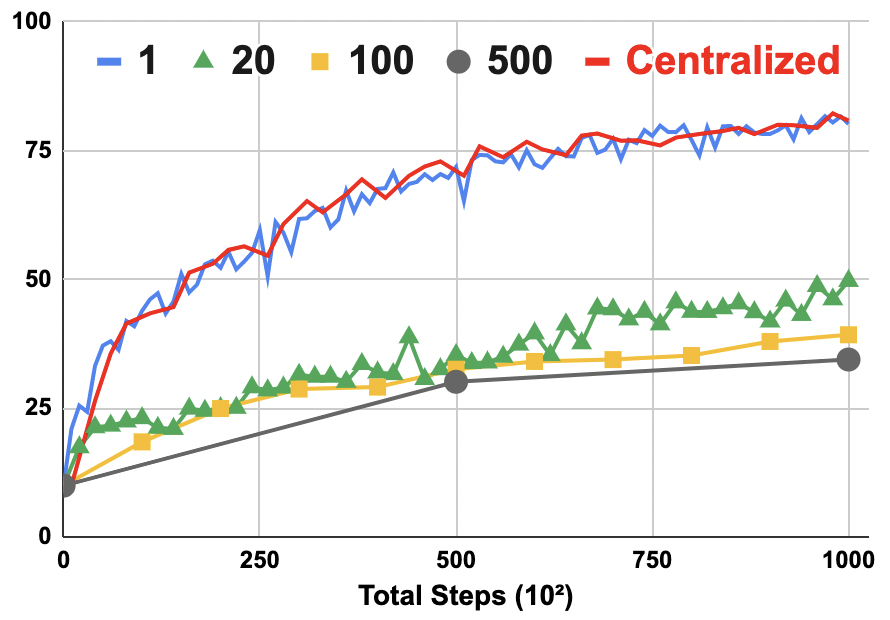}
    \vskip-5pt
    \mbox{\small (a) \GN}
    \endminipage
    \hfill
    \minipage{0.5\linewidth}
    \centering
    \includegraphics[width=1\linewidth]{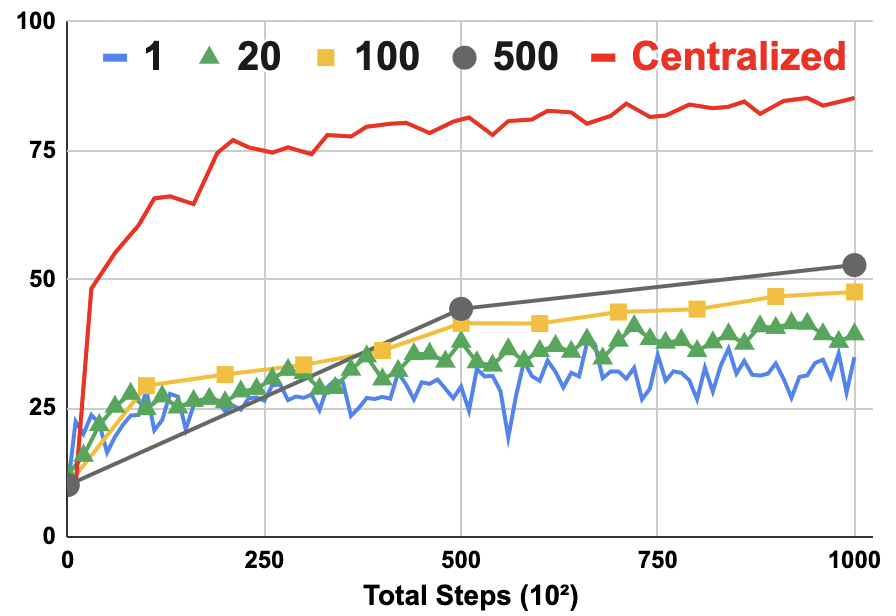} 
    \vskip-5pt
    \mbox{\small (b) \BN}    
    \endminipage
    \vspace{-5pt}
    \caption{\small Test accuracy on CIFAR-10 for different local steps ($E$) per round, given a fixed total SGD steps. X-axis: training progress.}
    \label{fig:fix_steps}
\end{figure}

In~\autoref{fig:fix_steps}, we conduct a finer-grained analysis of the opposite trend. Starting from the same initial model, we vary the number of local SGD steps per round but fix the total SGD steps. We show the test accuracy of the (aggregated) global model as the training proceeds (x-axis).
Comparing \BN and \GN, we see a drastically different effect of $E$. In particular, while the performance of \GN drops along with increasing $E$, \BN somehow benefits from a larger $E$. Namely, \BN seems unreasonably robust to training with fewer communication rounds but more local steps per round, suggesting the need for a deeper analysis of \BN in FL. 

\section{Conclusion}
\label{sec:diss}

We revisit the use of \BN and its common alternative, \GN, in federated learning (FL) and conduct an in-depth analysis. We find \BN outperforms \GN in many FL settings, except for high-frequency and extreme non-IID regimes. We reinvestigate the issues with \BN in FL and propose a simple yet effective practice, \Ours, to mitigate them while retaining the positive impacts of \BN. We hope our study provides the community with a good foundation for the full (theoretical) understanding of \BN towards training deeper models in FL. 

\newpage

\section*{Acknowledgments}
\label{suppl-sec:ack}
This research is supported in part by grants from the National Science Foundation (IIS-2107077 and OAC-2112606) and Cisco Research. We are thankful for the generous support of the computational resources by the Ohio Supercomputer Center.

\bibliographystyle{icml2024}
\bibliography{main}

\newpage
\onecolumn
\appendix
\begin{center}
\textbf{\LARGE Appendix}
\end{center}
\renewcommand{\thesection}{\Alph{section}}
\renewcommand{\thetable}{\Alph{table}}
\renewcommand{\thefigure}{\Alph{figure}}
\renewcommand{\theequation}{\Alph{equation}}

We provide details omitted in the main paper. 
\begin{itemize}
    \item \autoref{suppl-sec:exp_s}: details of experimental setups (cf.~\autoref{sec:method} and~\autoref{sec:study} of the main paper).
    \item \autoref{suppl-sec:exp_r}: additional experimental results and analysis (cf.~\autoref{sec:method} and~\autoref{sec:study} of the main paper). 
    \item \autoref{suppl-sec:algo}: detailed algorithm implementation of \Ours (cf. \autoref{sec:method} of the main paper).
    \item \autoref{suppl-sec:curves}: training curves for \Ours vs. BN vs. GN considering different non-IID partition and local steps.
\end{itemize}

\begin{table*}[h]
    \caption{Summary of datasets and setups.}
    \centering
    \footnotesize
    \setlength{\tabcolsep}{1.5pt}
	\renewcommand{\arraystretch}{1}
    \resizebox{\textwidth}{!}{
    \begin{tabular}{l|cccccccc}
    \toprule
    Dataset & Task & $\#$Class & $\#$Training & $\#$Test/Valid & $\#$Clients & Resolution & Networks\\
    \midrule
    CIFAR-10 & Classification & 10 & $50K$ & $10K$  & $5\sim 100$ & $32^2$ & LeNet-CNN, ResNet-20\\
    \midrule
    Tiny-ImageNet & Classification & 200 & $100K$ & $10K$ & $10$ & $64^2$ & ResNet-18\\
    \midrule
    ImageNet & Classification & 1,000 & $1,200K$ & $100K$ & $100$ & $224^2$ & ResNet-18\\
    \midrule
    Cityscapes & Segmentation & 19& $3K$ & $0.5K$  & $18$ & $768^2$& 
    \makecell[c]{DeepLabv3 + \\ \{MobileNet-v2, ResNet-50\}}\\
    \bottomrule
    \end{tabular}
    }
    \label{sup-tbl:dataset}
\end{table*}

\begin{table*}[h] 
    \caption{Default FL settings and training hyperparameters in the main paper.}
    \centering
    \footnotesize
    \setlength{\tabcolsep}{2pt}
	\renewcommand{\arraystretch}{1.25}
    \resizebox{\textwidth}{!}{
    \begin{tabular}{l|ccccccccc}
    \toprule
    Dataset & Non-IID & Sampling & Optimizer & Learning rate & Batch size & $T^\star$ for \Ours\\
    \midrule
    CIFAR-10 & \makecell[c]{Shards, \\ Dirichlet(\{0.1, 0.3, 0.6\}), \\ IID} & $10\sim100\%$ & \makecell[c]{SGD + \\0.9 momentum} & 0.2/0.02 & 20 & $50\%$ of total rounds\\
    \midrule
    Tiny-ImageNet & \makecell[c]{Shards, \\ Dirichlet(\{0.1, 0.3, 0.6\}), \\ IID} & 50\% & \makecell[c]{SGD + \\0.9 momentum} & 0.02 & 20 & $50\%$ of total rounds\\
    \midrule
    ImageNet & Dirichlet 0.1 & $10\%$ & \makecell[c]{SGD + \\0.9 momentum}  & 0.1 & 20 & $50\%$ of total rounds\\ 
    \midrule
    Cityscapes & Cities & 50\% & Adam & 0.01/0.001 & 8 & 90th round\\
    \bottomrule
    \end{tabular}
    }
    \label{sup-tbl:hyper}
\end{table*}

\section{Experiment Details}
\label{suppl-sec:exp_s}

\subsection{Datasets, FL settings, and hyperparameters}
We use \FedAvg for our studies, with weight decay $1\mathrm{e}{-4}$ for local training. Learning rates are decayed by $0.1$ at $50\%, 75\%$ of the total rounds, respectively. Besides that, we summarize the training hyperparameters for each of the federated experiments included in the main paper in~\autoref{sup-tbl:hyper}. Additionally, for the Cityscape experiments in~\autoref{tbl:cityscapes}, we make each ``city'' a client and run $100$ rounds, with local steps to be $5$ epochs. More details about the datasets are provided in~\autoref{sup-tbl:dataset}.

For pre-processing, we generally follow the standard practice which normalizes the images and applies some augmentations. CIFAR-10 images are padded $2$ pixels on each side, randomly flipped horizontally, and then randomly cropped back to $32 \times 32$. For Tiny-ImageNet, we simply randomly cropped to the desired sizes and flipped horizontally following the official PyTorch ImageNet training script. For the Cityscapes dataset, we use output stride $16$. In training, the images are randomly cropped to $768\times768$ and resized to $2048\times1024$ in testing.

\section{Additional Experimental Results and Analysis}
\label{suppl-sec:exp_r}

\subsection{Additional study of fixing \BN parameters}
In~\autoref{ss:mismatch}, we discuss that the \BN statistics are the main critical parameters in FL and thus motivate our design in \Ours to fix the \BN statistics to be the global aggregated ones after certain rounds. Here we include a further study to confirm the importance of \BN statistics by comparing them with the learnable affine transformation parameterized by $(\vgamma, \vbeta)$. 

For \Ours, besides fixing the \BN statistics from round $T^\star$, we consider fixing the $(\vgamma, \vbeta)$ alone or together. The results on CIFAR-10 (Shards, fixed epochs, $E = 100$) setting using ResNet20 is in~\autoref{sup-tab:ablation}. We observe that fixing the $(\vgamma, \vbeta)$ only has slight effects on the test accuracy either in combination with fixing $(\vgamma, \vbeta)$ or not, validating that the statistics are the main reason making it suffers more in FL, compared to the affine transformation. Fixing $(\vgamma, \vbeta)$ alone cannot match the performance of the originally proposed \Ours.

\begin{table*}[h]
\caption{\small \textbf{Fixing different parameters as in \Ours.} We consider fixing the \BN statistics $(\vmu, \vsigma)$ as in original \Ours or fixing the parameters $(\vgamma, \vbeta)$ of the affine transformation in \BN layers.
on CIFAR-10 (Shards, fixed epochs, $E = 100$) setting using ResNet20.}
\label{sup-tab:ablation}
\centering
\begin{tabular}{lc|c}
\toprule
$(\vmu, \vsigma)$ & $(\vgamma, \vbeta)$ & Acc (\%)\\
\midrule
\cmark & \cmark  & 75.22 \\
\cmark & \xmark  & 76.56 \\
\xmark & \cmark  & 55.33 \\
\xmark & \xmark  & 53.97 \\
\bottomrule
\end{tabular}
\end{table*}

\subsection{Additional figure for maintained SGD momentum on Tiny-ImageNet in~\autoref{ss:ind}}
In~\autoref{fig:fix_mom}, we show the effect of maintained SGD momentum to \FedAvg on CIFAR-10 and provide analysis in~\autoref{ss:ind}. Here we show that the same effect is observed on Tiny-ImageNet. 

\begin{figure}[h]
    \vskip 0pt
    \hfill
    \minipage{1\linewidth}
·   \minipage{1\linewidth}
    \centering
    \includegraphics[width=0.4\linewidth]{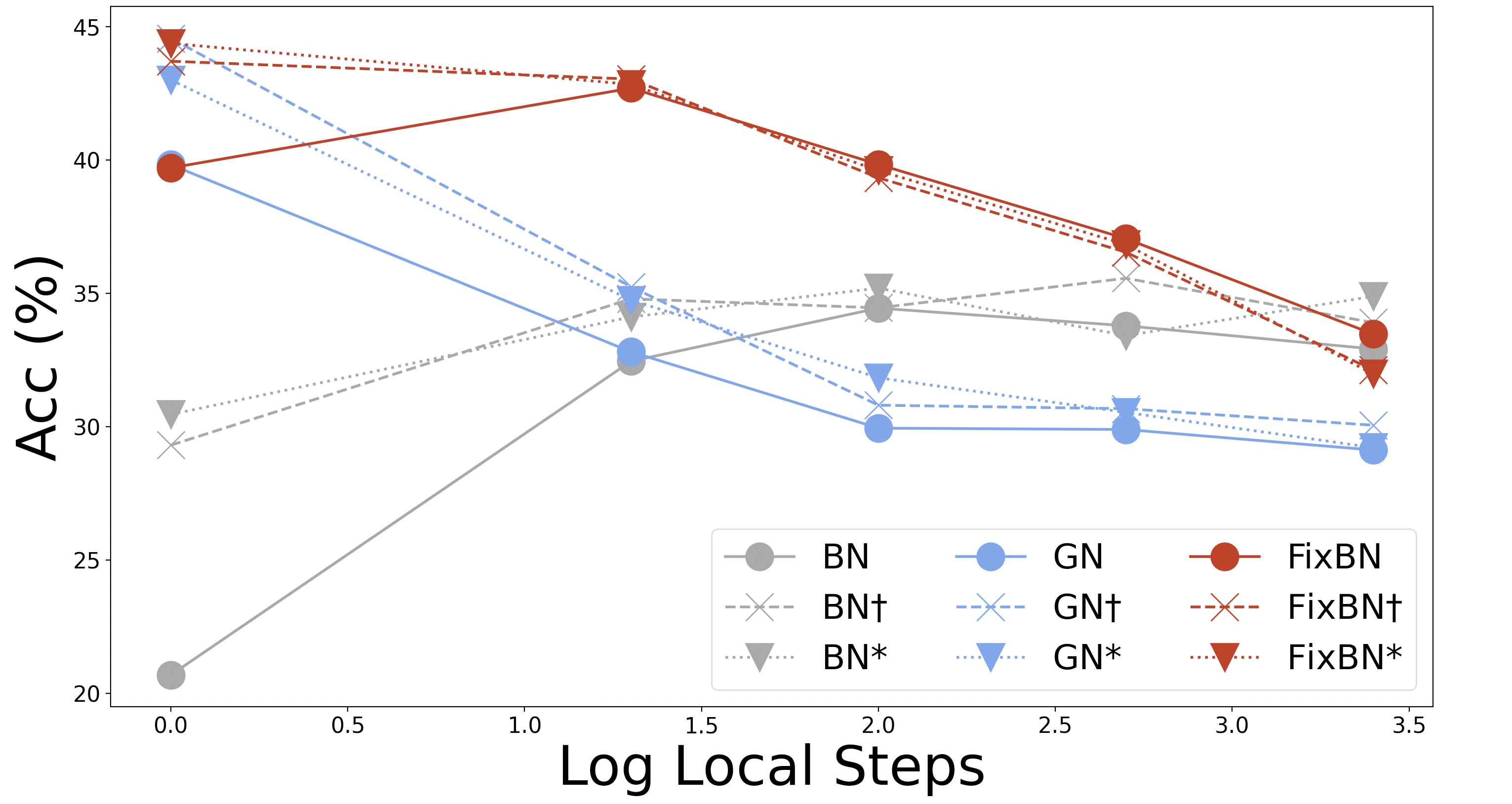}
    \endminipage
    \endminipage
    \vspace{-8pt}
    \caption{\small \textbf{Maintained momentum results on Tiny-ImageNet.} We apply maintained \textbf{global momentum ($^\dagger$)} and \textbf{local momentum ($^\star$)} to \FedAvg with different normalizers. The setting is (Shards, fixed 128 epochs) with different local steps $E$. Please refer to \autoref{fig:fix_mom} in the main paper for results on Cifar-10.}
    \vspace{-14pt}
    \label{fig:fix_mom_tiny}
\end{figure}

\subsection{Additional figures for the empirical study in~\autoref{sec:study_new}}
In~\autoref{sec:study_new}, we provide a detailed empirical study to compare \BN and \GN across various FL settings to understand their sweet spots. We provide a closer look at the observations we summarized in the main paper.

\begin{figure}[H]
    \centering
    \minipage{0.6\linewidth}
    \minipage{0.5\linewidth}
    \centering
    \vskip-5pt
    \includegraphics[width=1\linewidth]{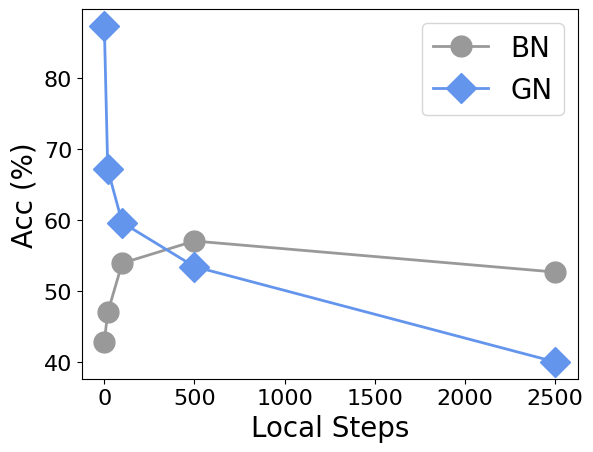} 
    \mbox{\small (a) CIFAR-10}
    \endminipage
    \hfill
    \minipage{0.5\linewidth}
    \centering
    \vskip-5pt
    \includegraphics[width=1\linewidth]{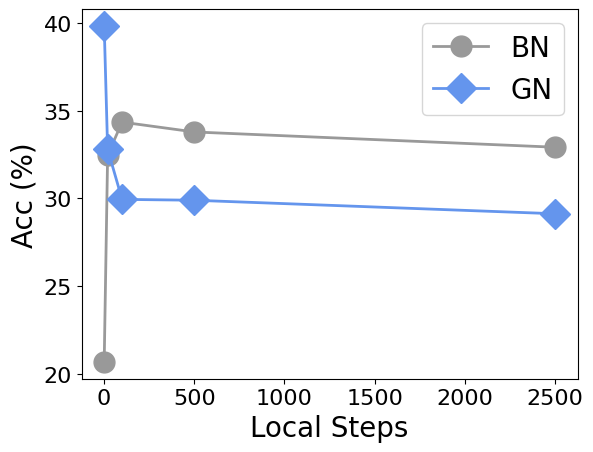}
    \mbox{\small (b) Tiny-ImageNet}
    \endminipage
    \endminipage
    \vskip -5pt
    \caption{\small \textbf{The opposite trends along \#local steps $E$.} We consider the (Shards, \textbf{fixed epochs}) setting: the more the local step $E$ is, the fewer the total number of communication rounds is.
    \GN drops with less communication as expected, while \BN can improve.}
    \label{fig:trend}
    \vskip -10pt
\end{figure}

\begin{figure}[H]
    \centering
    \minipage{0.6\linewidth}
    \minipage{0.5\linewidth}
    \centering
    \includegraphics[width=1\linewidth]{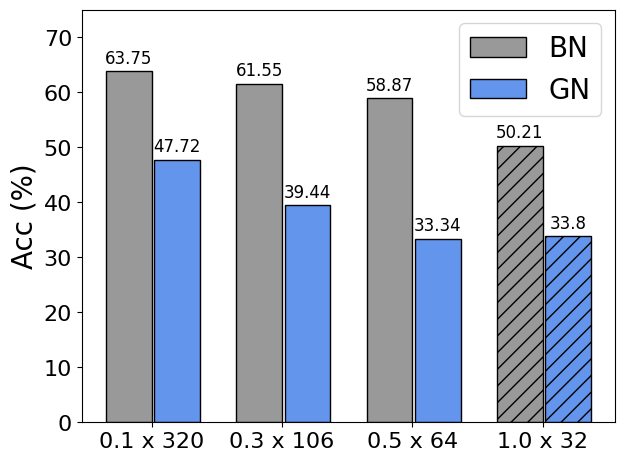} 
    \mbox{\small (a) Participation (\%) $\times$ \#Rounds}
    \endminipage
    \hfill
    \minipage{0.5\linewidth}
    \centering
    \vskip-5pt
    \includegraphics[width=1\linewidth]{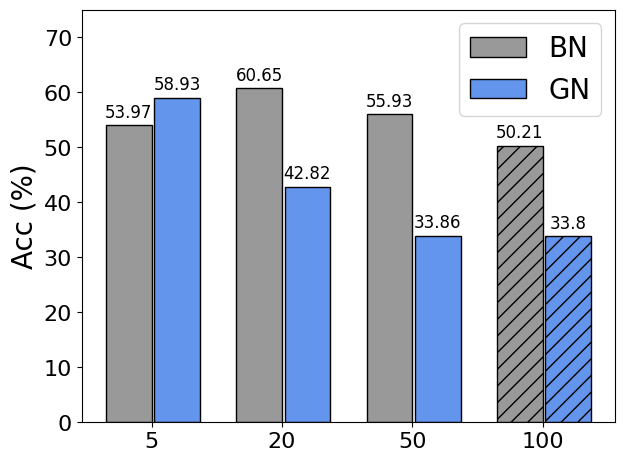}
    \mbox{\small (b) Different \#clients}
    \endminipage
    \endminipage
    \caption{\small \textbf{More settings.} We consider more clients ($M=5\sim100, E=100$) for partitioning CIFAR-10 (Shards) with fixed epochs and varying the participation rate of clients every round.}
    \label{fig:factors}
    \vskip -10pt
\end{figure}

\begin{itemize}
\item \textbf{The trends along the number of local steps $E$ per communication round.} In~\autoref{fig:fix_steps}, we identify the opposite trends along \#local steps $E$ between \BN and \GN. Here we provide an expanded view. As shown in~\autoref{fig:trend}, we see GN drops with less communication as expected due to the well-known non-IID model drift problem in FL. Interestingly, we found that BN can actually improve within a certain range of communication frequencies (for local steps in [1,500]), which suggests that further investigation and theoretical analysis are required for \BN in FL.

\item \textbf{More settings.} We further verify that factors such as participation rate and the number of clients for partitioning the data in ~\autoref{fig:factors}. As expected, the results are consistent with the observations summarized in~\autoref{sec:study_new}, particularly in that there is no definite winner between BN and GN while BN often outperforms GN.
\end{itemize}

\subsection{Different \# of groups for GN}
For experiments in our study, we set the $\#$ of groups $=2$ for \GN layers. We did not find the group size a significant factor for the performance, as confirmed in~\autoref{tab:gsize}.

\begin{table*}[h]
\caption{\small \textbf{Effects of the groupsize for \GN.} We experiment with different \# of groups ($2\sim8$) to divide the channels in \GN layers in the CIFAR-10 (Shards, $E=100$) with fixed epochs setting.}
\label{tab:gsize}
\centering
\begin{tabular}{c|c}
\toprule
Groupsize & Acc(\%) \\

\midrule
$2$ & 59.42\\
$4$ & 57.61\\
$8$ & 58.86\\
\bottomrule
\end{tabular}
\end{table*}

\subsection{Effects of batch size for \BN} We experiment with various batch sizes for both BN and \Ours in the CIFAR-10 (Shards, $E=1$) setting and saw \Ours maintains the advantage over standard \FedAvg+BN.

\begin{figure}[H]
    \centering

    \vskip-5pt
    \includegraphics[width=0.5\linewidth]{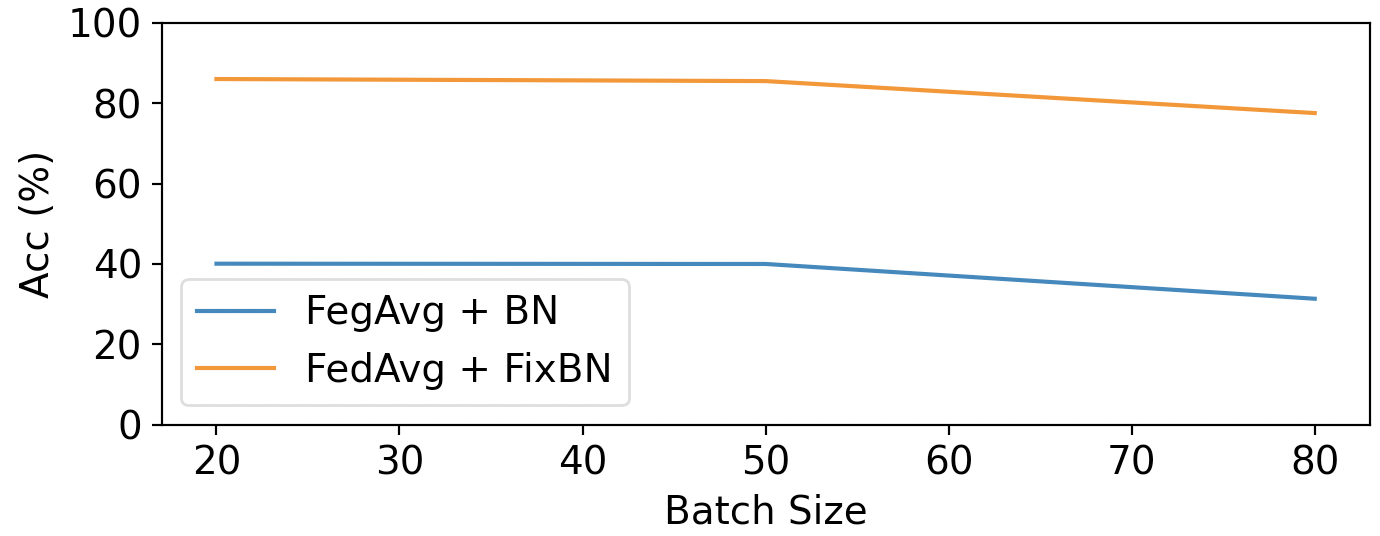}
    \caption{\small \Ours maintains advantage over different batch size selections.}
\end{figure}

\subsection{Method for selecting $T^\star$}
\label{suppl-sec: indicator}
$T^\star$ denotes the round for which we begin to ``freeze" \BN layer in local training. As discussed in \autoref{ss_insight} and shown in \autoref{fig:fixbn_motiv} (b) and (c), what enables the replacement of stochastically calculated local statistics in local training with fixed statistics is the reduction of variation in local statistics and convergence of local models. While these observations do not give a definitive answer to exactly where $T^\star$ exists, they lead to three natural approaches: 1) assuming the training budget allows, the user could observe local mini-batch statistics in train time and visually determine a good $T^\star$ for their specific tasks in which the local mini-batch statistics or loss converge. 2) Alternatively, one could adapt a sliding window and decide based on the rate of change of local statistics within this window. Specifically, define W as the size of a sliding window, define $\tau$ as the FixBN threshold, compute $\text{Var}_{\Delta m,t} = \frac{1}{W} \sum_{i=t-W+1}^{t} (\Delta \text{BN}_{m,i} - \overline{\Delta \text{BN}_{m,t}})^2$, if this falls under $\tau$, then we apply FixBN. With this setup, we recommend using a large W and small $\tau$ to account for instability during training. 3) Additionally, it could be done by cross-validation, or we can apply federated hyperparameter optimization proposed in \citet{khodak2021federated}, similar to the general approach to determine a hyperparameter.

\section{\Ours Algorithm}
\label{suppl-sec:algo}
\begin{algorithm}[H]
\SetAlgoLined
\caption{\Ours: Federated Learning with Fixed Batch Statistics}
\label{alg:fedbasis}
\SetKwInOut{Input}{Input}
\SetKwInOut{SInput}{Server input}
\SetKwInOut{SOutput}{Server output}
\SetKwInOut{CInput}{Client $m$'s input}
\SInput{initial global model weights $\bar{\vtheta}_{1}$, fixing round $T^{\star}$ (in~\autoref{ss_two_stage}), total number of rounds $T$}
\For{$t\leftarrow 1$ \KwTo $T$ rounds}{
    \textbf{Communicate} $\bar{\vtheta}_{t}$ to all (sampled) clients $m\in [M]$;\\
    \For{each client $m\in [M]$ in parallel}{
        \If{$t > T^{\star}$}{
            \Indp\For{each BN layer $\{f_{BN}^l{(\bar{\vtheta}_{t})}\}$ parametrized by $\bar{\vtheta}_{t}$}{
         $f_{BN}^l({\bar{\vtheta}_{t}})$.eval(); \tcp{set into eval mode to fix BN statistics}
        }
        }
        $\vtheta_{t+1}^{m} \gets \text{ClientUpdate}(m, \bar{\vtheta}_{t})$; \tcp{normal client updates}
        \textbf{Communicate} ${\vtheta}_{t+1}^{m}$ to the server;
    }
    \textbf{Construct} $\bar{\vtheta}_{t+1} = \frac{1}{M}\sum_{m=1}^M {\vtheta}_{t+1}^{m}$;
}
\SOutput{$\bar{\vtheta}_{T+1}$}
\end{algorithm}

\section{Training Curves}
\label{suppl-sec:curves}
\begin{figure}[H]
    \centering
    \minipage{1\linewidth}
    
    \centering
    \vskip-5pt
    \includegraphics[width=0.8\linewidth]{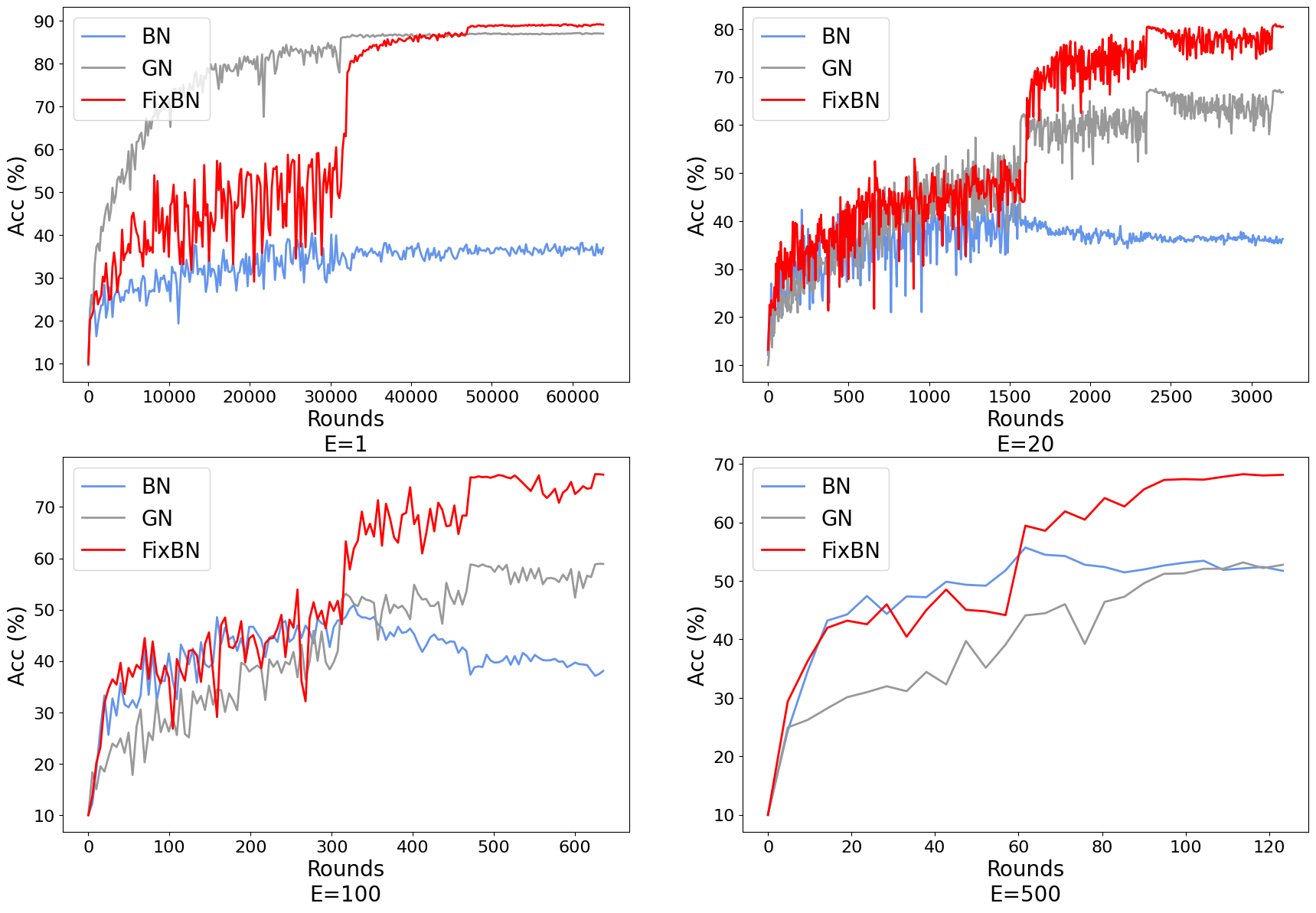} 
    \endminipage
    
    \caption{\small Convergence curves of the test accuracy of CIFAR-10 in the fixed epoch, shards non-IID partitions, and \textbf{$E=1\sim500$} setting.}
    \label{sup-fig:shards}
    \vskip -5pt
\end{figure}

\begin{figure}[H]
    \centering
    \minipage{1\linewidth}
    
    \centering
    \vskip-5pt
    \includegraphics[width=0.8\linewidth]{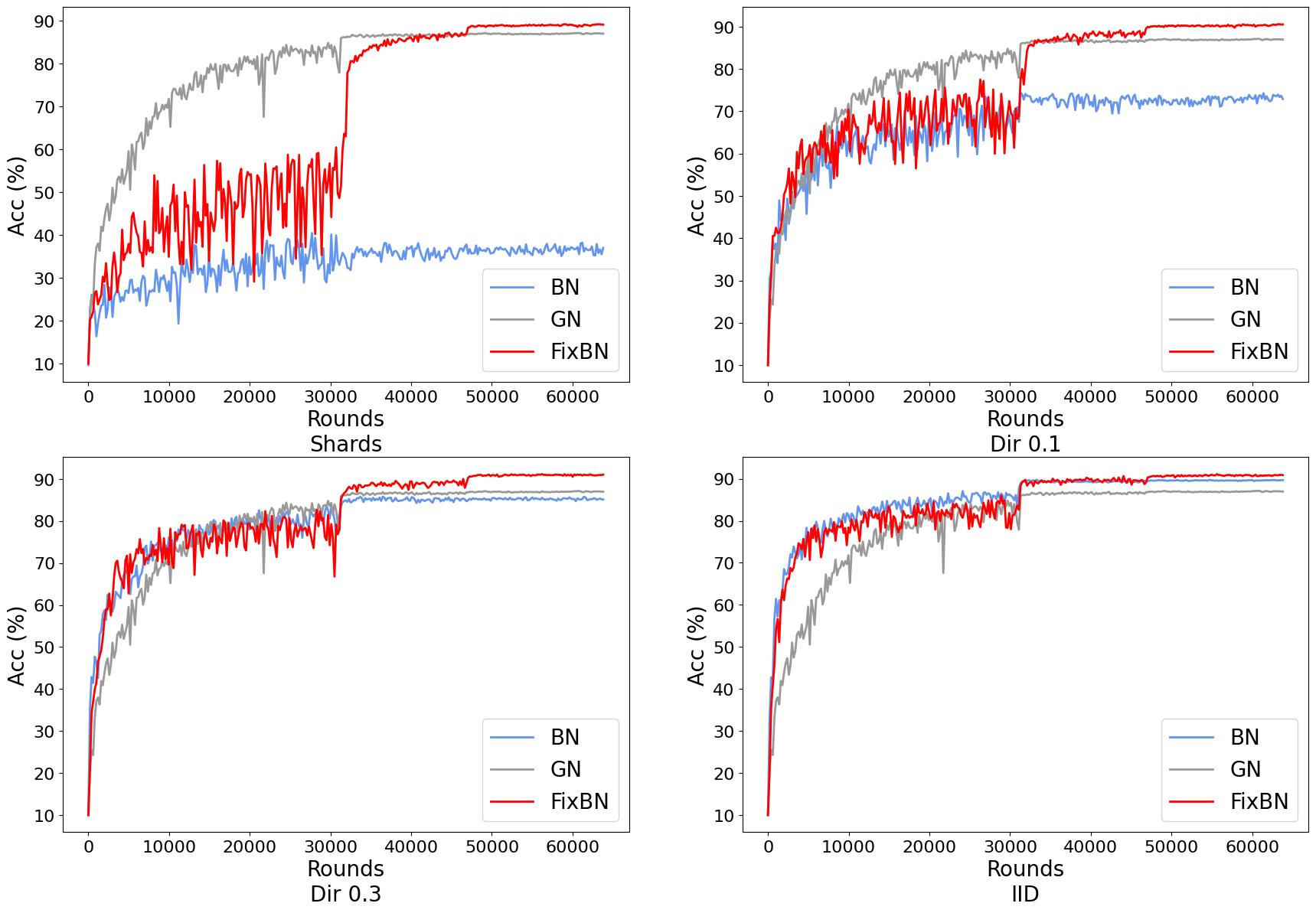} 
    \endminipage
    
    \caption{\small Convergence curves of the test accuracy of CIFAR-10 in the fixed epoch, \textbf{different non-IID partitions}, and \textbf{$E=1$} setting.}
    \label{sup-fig:s1}
    \vskip -5pt
\end{figure}

\begin{figure}[H]
    \centering
    \minipage{1\linewidth}
    
    \centering
    \vskip-5pt
    \includegraphics[width=0.8\linewidth]{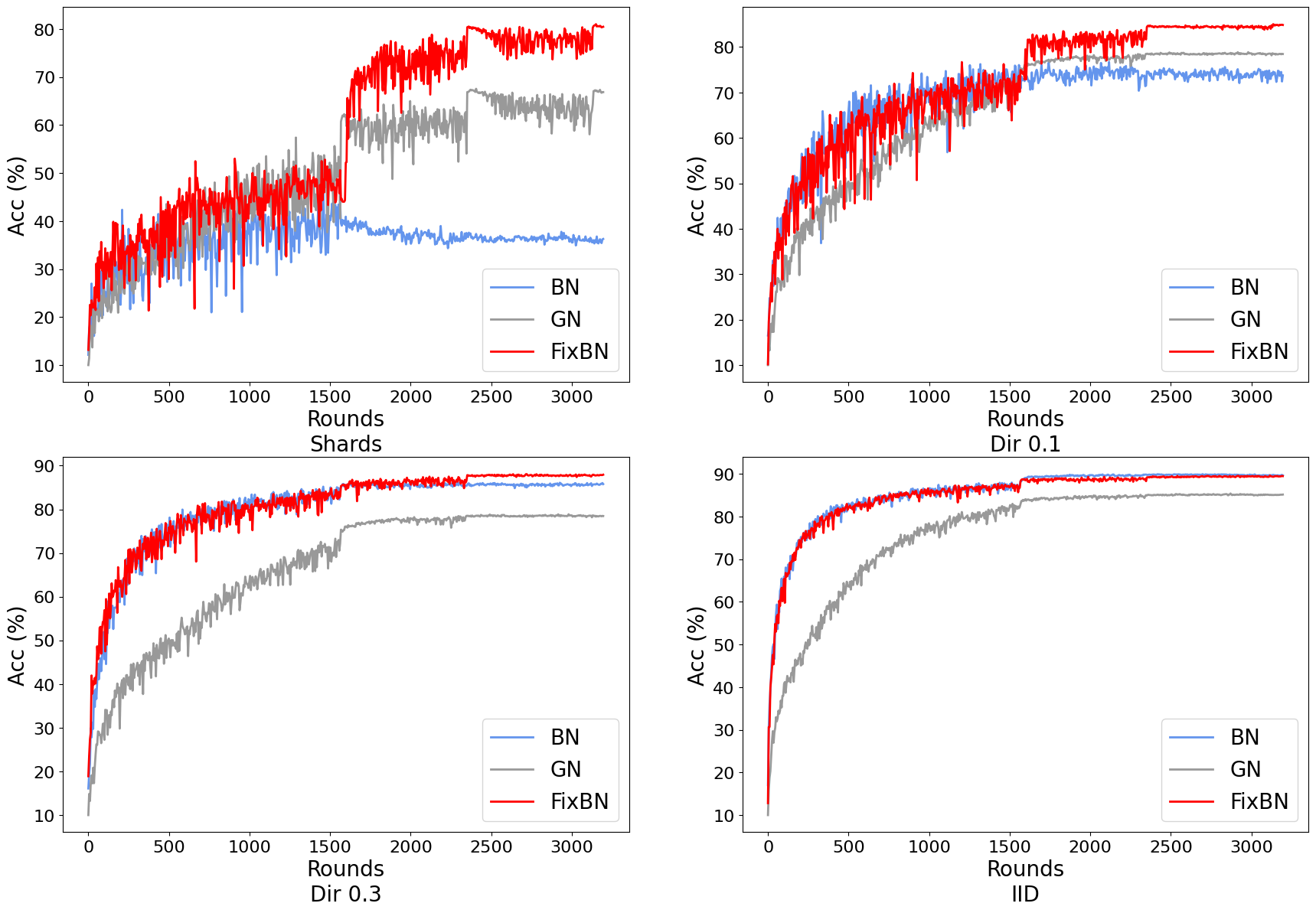} 
    
    \endminipage
    
    \caption{\small Convergence curves of the test accuracy of CIFAR-10 in the fixed epoch, \textbf{different non-IID partitions}, and \textbf{$E=20$} setting.}
    \label{sup-fig:s20}
    \vskip -5pt
\end{figure}

\begin{figure}[H]
    \centering
    \minipage{1\linewidth}
    
    \centering
    \vskip-5pt
    \includegraphics[width=0.8\linewidth]{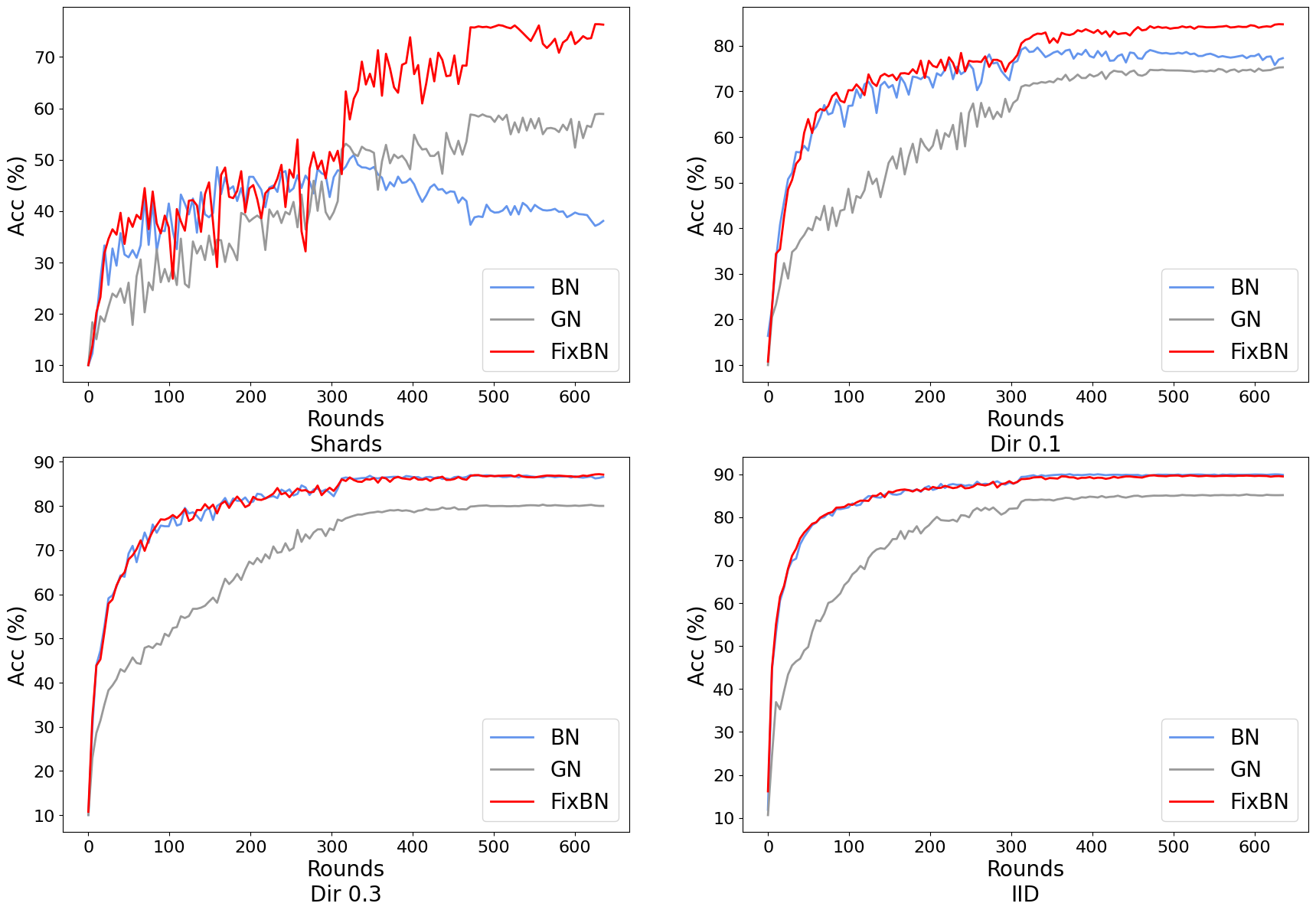} 
   
    \endminipage
    
    \caption{\small Convergence curves of the test accuracy of CIFAR-10 in the fixed epoch, \textbf{different non-IID partitions}, and \textbf{$E=100$} setting.}
    \label{sup-fig:s100}
    \vskip -5pt
\end{figure}

We provide the training curves of \Ours and other normalizers under various settings with fixed $128$ epochs using ResNet20 in ~\autoref{sup-fig:shards},~\autoref{sup-fig:s1},~\autoref{sup-fig:s20}, and~\autoref{sup-fig:s100}, corresponding to~\autoref{sec:study}. 

\end{document}